\documentclass[lettersize,journal]{IEEEtran}
\usepackage{amsmath,amsfonts}
\usepackage{algorithmic}
\usepackage{algorithm}
\usepackage{array}
\usepackage[caption=false,font=normalsize,labelfont=sf,textfont=sf]{subfig}
\usepackage{textcomp}
\usepackage{stfloats}
\usepackage{url}
\usepackage{verbatim}
\usepackage{graphicx}
\usepackage{cite}
\usepackage{algorithm}
\usepackage{algorithmic}
\usepackage{xcolor,soul}
\usepackage{booktabs}
\usepackage{url}
\usepackage{graphicx}
\usepackage{amsthm}
\usepackage{amssymb}
\usepackage{amsfonts}
\usepackage{amsmath}
\usepackage{multirow}
\newcommand{\eat}[1]{}
\newcommand{\etal}{{et al.~}}       
\newcommand{\eg}{{e.g.~}}           
\newcommand{\ie}{{i.e.~}}           
\newcommand{\wrt}{{w.r.t.~}}         
\newcommand{\aka}{{a.k.a.~}}        

\newtheorem{definition}{Definition}
\newtheorem{problem}{Problem}

\newtheorem{theorem}{Theorem}
\makeatletter
\newcommand*{\indep}{%
	\mathbin{%
		\mathpalette{\@indep}{}%
	}%
}
\newcommand*{\nindep}{%
	\mathbin{
		\mathpalette{\@indep}{\not}
	}%
}
\newcommand*{\@indep}[2]{%
	\sbox0{$#1\perp\m@th$}
	\sbox2{$#1=$}
	\sbox4{$#1\vcenter{}$}
	\rlap{\copy0}
	\dimen@=\dimexpr\ht2-\ht4-.2pt\relax
	\kern\dimen@
	{#2}%
	\kern\dimen@
	\copy0 
}

\newcommand\ci{\perp\!\!\!\perp}

\begin{document}

\title{Disentangled Representation Learning for Causal Inference with Instruments}

\author{Debo Cheng, Jiuyong Li, Lin Liu, Ziqi Xu, Weijia Zhang, Jixue Liu
	and Thuc Duy Le 
\thanks{D. Cheng, J. Li, L. Liu, J. Liu and T. Le are with UniSA STEM, University of South Australia, Adelaide, SA, 5095, Australia. E-mail: chedy055@mymail.unisa.edu.au, \{Jiuyong.Li, Lin.Liu. Jixue.Liu, Thuc.Le\}@unisa.edu.au. (Corresponding author: Debo Cheng)

	Z. Xu, is with School of Computing Technologies, RMIT University, Melbourne, 3000, Australia. E-mail: ziqi.xu@rmit.edu.au.
	
	W. Zhang is with School of Information and Physical Sciences, University of Newcastle, Callaghan, NSW, 2308, Australia. E-mail: weijia.zhang@newcastle.edu.au.
	
	We wish to acknowledge the support from the Australian Research Council (under grant DP230101122). 
	
	}
 }




\maketitle

\begin{abstract}
	Latent confounders are a fundamental challenge for inferring causal effects from observational data. The instrumental variable (IV) approach is a practical way to address this challenge. Existing IV based estimators need a known IV  or other strong assumptions, such as the existence of two or more IVs in the system, which limits the application of the IV approach. In this paper, we consider a relaxed requirement, which assumes there is an IV proxy in the system without knowing which variable is the proxy. We propose a Variational AutoEncoder (VAE) based disentangled representation learning method to learn an IV representation from a dataset with latent confounders and then utilise the IV representation to obtain an unbiased estimation of the causal effect from the data. Extensive experiments on synthetic and real-world data have demonstrated that the proposed algorithm outperforms the existing IV based estimators and VAE-based estimators. 
\end{abstract}


\section{Introduction}
\IEEEPARstart{E}{stimating} the causal effect of a treatment (\aka intervention, or exposure) on an outcome is a fundamental task in many areas~\cite{pearl2009causality,hernan2010causal}, such as policy-making and new drug evaluation, etc. Randomised controlled trials (RCTs) are the gold standard for inferring causal effects, but they are often impractical in real-world applications due to time or ethical constraints~\cite{chalmers1981method}. Thus, causal effect estimation with observational data has become an alternative to RCTs.

\begin{figure}[t]
	\centering
	\includegraphics[scale=0.4]{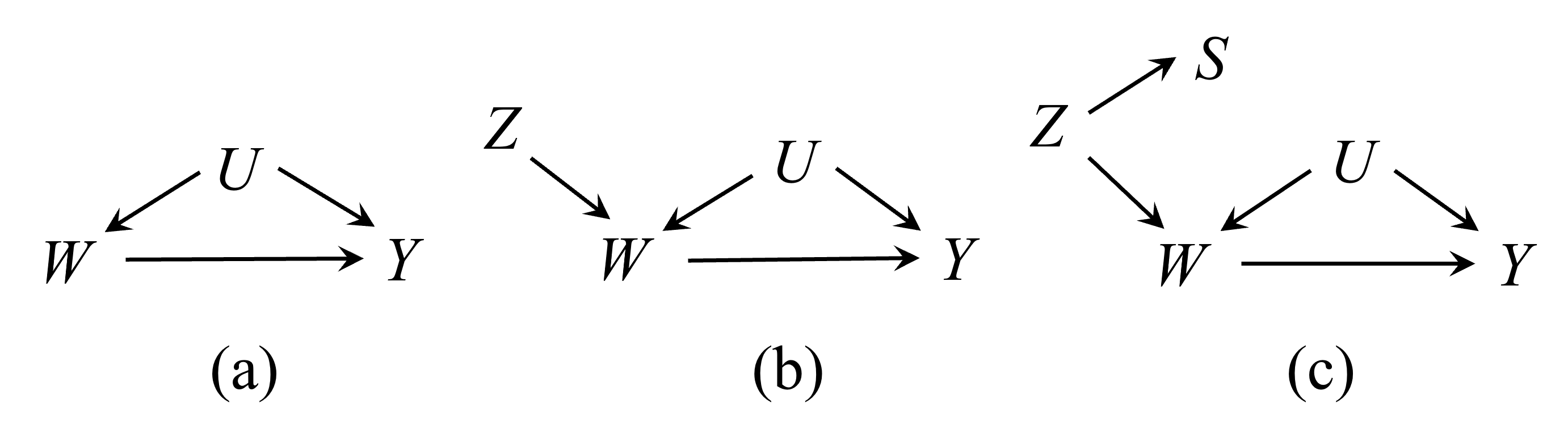}
	\caption{Causal  graphs with latent variables to show the problem of causal effect estimation from observational data. In DAG (a), the causal effect of $W$ on $Y$ cannot be estimated from observational data; in DAG (b), there is a valid IV $Z$; in DAG (c), $Z$ is an unmeasured IV and $S$ is a  surrogate IV (SIV) of $Z$. The causal effect of $W$ on $Y$ in both DAGs (b) and (c) can be recovered from observational data. }
	\label{fig:exp001}
\end{figure}

However, estimating causal effects using observational data suffers from confounding bias, due to the spurious association caused by confounders  that affect  both the treatment and outcome variables. Unmeasured confounders make the situation even worse~\cite{spirtes2000causation,chen2021causal,cheng2022toward}. As shown in Fig.~\ref{fig:exp001} (a), if there is an unmeasured confounder $(U)$ between the treatment $(W)$ and the outcome $(Y)$, the causal effect of $W$ on $Y$ cannot be estimated with observational data except there is an instrumental variable~\cite{hernan2006instruments,pearl2009causality,imbens2015causal}.  

The instrumental variable (IV) approach is a commonly used way to estimate causal effects from data when the unconfoundedness assumption is violated~\cite{angrist1995two,pearl2009causality}. A valid IV (denoted as $Z$) must satisfy the following three conditions~\cite{brito2002generalized,hernan2006instruments,sjolander2019instrumental}: (i) $Z$ influences the treatment (\ie \textit{relevance condition}), (ii)  the causal effect of $Z$ on $Y$ is only through $W$ (\aka \textit{exclusion restriction}), and (iii) $Z$ and $Y$ do not have any common causes (\ie \textit{unconfounded instrument}). The three conditions of a valid IV can only be verified using domain knowledge or based on the underlying causal graph of the system but not from data~\cite{pearl1995testability}. It is known that domain knowledge of an IV or a causal graph is rarely available in many real applications~\cite{hernan2006instruments}. 
Therefore, it is desirable to explore an effective data-driven method to discover {a valid IV} directly from data.

A few data-driven IV based methods have been proposed for causal effect estimation without assuming a known IV, but they often have other constraints. For example, IV.Tetrad~\cite{silva2017learning} requires that at least a pair of IVs exist in the system and the set of all the remaining variables excluding the pair of IVs is a conditioning set  of the IV with respect to the treatment and the outcome. sisVIVE~\cite{kang2016instrumental} requires that at least half of the variables (the set of candidate IVs) are valid IVs. 

Some research has proposed the necessary conditions of IVs for obtaining a bound estimation of a causal effect, i.e. a multiset of possible estimates from data, instead of a unique estimate. For instance, Pearl~\cite{pearl1995testability} proposed an instrumental inequality to find a set of candidate IVs from data with discrete variables, and Kuroki and Cai extended the instrumental inequality to linear structural equation models with continuous variables~\cite{kuroki2005instrumental}. Assuming a linear non-Gaussian acyclic causal model, Xie et al.~\cite{xie2022testability} proposed a necessary condition based on a generalised independent noise condition for identifying continuous variables as the candidate valid IVs, but the condition only produces a  bound estimation. 

\IEEEpubidadjcol

Therefore, the existing data-driven IV methods either rely on strong assumptions, or only provide a necessary condition for determining candidate IVs. In order to develop a more effective and practical data-driven IV based causal effect estimator, in this paper, we consider a relaxed requirement, which assumes there exists at least one IV proxy (a.k.a. surrogate IV, \ie SIV) in the system without knowing which variable is the proxy. The assumption of an IV proxy is practical since there exist many proxy variables for latent confounders~\cite{montgomery2000measuring,miao2018identifying}. 

It is challenging to determine IVs (or SIVs) from the set of measured covariates since IVs (or SIVs) and measured confounders are statistically inseparable. To address this challenge, we propose a data-driven method, \underline{D}isentangled \underline{IV} based on \underline{VAE} (DIV.VAE)  using disentangling techniques~\cite{hassanpour2019learning,zhang2021treatment} to learn the latent representation $\mathbf{\Phi}$ of the set of pretreatment variables  $\mathbf{X}$, which are measured before applying the treatment $W$ and observing the outcome $Y$~\cite{imbens2015causal},  and disentangle $\mathbf{\Phi}$ into $(\mathbf{Z}, \mathbf{C})$, where $\mathbf{Z}$ represents IV and $\mathbf{C}$ represents confounders in the latent space. To the best of our knowledge,  DIV.VAE is the first work using the VAE model to infer the IV representation from observed pretreatment variables when the unconfoundedness assumption is not satisfied.

Our main contributions are summarised as follows.  

\begin{itemize}
	\item We address the challenging problem of causal effect estimation from data in the presence of latent confounders.
	\item We propose a novel disentangled representation learning method, DIV.VAE to learn the latent IV representation and the latent confounding representation for achieving unbiased causal effect estimation. 
	\item  We empirically evaluate the effectiveness of DIV.VAE on synthetic and real-world datasets, in comparison with state-of-the-art causal effect estimators. The results show that the DIV.VAE outperforms baseline estimators.
\end{itemize}

\section{Preliminaries}
\label{sec:Pre}
\subsection{Notations}
\label{subsec:notations}
We represent variables and their values with uppercase and lowercase letters, respectively. A set of variables and a value assignment of the set are denoted by bold-faced uppercase and lowercase letters, respectively. 

Let $\mathcal{G}\!\!=\!\!(\mathbf{V}, \mathbf{E})$ be a directed acyclic graph (DAG) where $\mathbf{V}\!\!=\!\!\{V_{1}, \dots, V_{p}\}$ are the set of nodes representing $p$ random variables and  $\mathbf{E} \!\subseteq \mathbf{V} \times \mathbf{V}$ are the set of edges representing the relationships between nodes. In DAG $\mathcal{G}$, two nodes are \textit{adjacent} when there exists a directed edge $\rightarrow$ between them. A path $\pi$ from $V_i$ to $V_j$ is a directed or causal path if all edges along it are directed towards $V_j$. If there is a directed path $\pi$ from $V_i$ to $V_j$, $V_i$ is known as an ancestor of $V_j$ and $V_j$ is a descendant of $V_i$. The sets of ancestors and descendants of a node $V$ are denoted as $An(V)$ and $De(V)$, respectively. 

A DAG is causal if the directed edge $V_i \rightarrow V_j$ between $V_i$ and $V_j$ indicates that $V_i$ is a direct cause of $V_j$. In a DAG $\mathcal{G}$, a path $\pi$ between $V_{i}$ and $V_{j}$ comprises a sequence of distinct nodes $\langle V_{i}, \dots, V_{j}\rangle$ with every pair of successive nodes being adjacent, and $V_i$ and $V_j$ are end nodes of $\pi$.

The definitions of Markov property and faithfulness are introduced in the following.
\begin{definition}[Markov property~\cite{pearl2009causality}]
	\label{Markov condition}
	Given a DAG $\mathcal{G}=(\mathbf{V}, \mathbf{E})$ and the joint probability distribution of $\mathbf{V}$ $(P(\mathbf{V}))$, $\mathcal{G}$ satisfies the Markov property if for $\forall V_i \in \mathbf{V}$, $V_i$ is probabilistically independent of all of its non-descendants, given the parent nodes of $V_i$.
\end{definition}

\begin{definition}[Faithfulness~\cite{spirtes2000causation}]
	\label{Faithfulness}
	A DAG $\mathcal{G}=(\mathbf{V}, \mathbf{E})$ is faithful to a joint distribution $P(\mathbf{V})$ over the set of variables $\mathbf{V}$ if and only if every independence present in $P(\mathbf{V})$ is entailed by $\mathcal{G}$ and satisfies the Markov property. A joint distribution $P(\mathbf{V})$ over the set of variables $\mathbf{V}$ is faithful to the DAG $\mathcal{G}$ if and only if the DAG $\mathcal{G}$ is faithful to the joint distribution $P(\mathbf{V})$.
\end{definition}

When the faithfulness assumption is satisfied between a joint distribution $P(\mathbf{V})$ and a DAG of a set of variables $\mathbf{V}$, the dependency/independency relations among the variables can be read from the DAG~\cite{pearl2009causality,spirtes2000causation}. In a DAG, d-separation is a well-known graphical criterion that is used to read off the  conditional independence between variables entailed in the DAG when the Markov property and faithfulness are satisfied~\cite{pearl2009causality,spirtes2000causation}.

\begin{definition}[d-separation~\cite{pearl2009causality}]
	\label{d-separation}
	A path $\pi$ in a DAG $\mathcal{G}=(\mathbf{V}, \mathbf{E})$ is said to be d-separated (or blocked) by a set of nodes $\mathbf{M}$ if and only if
	(i) $\pi$ contains a chain $V_i \rightarrow V_k \rightarrow V_j$ or a fork $V_i \leftarrow V_k \rightarrow V_j$ such that the middle node $V_k$ is in $\mathbf{M}$, or
	(ii) $\pi$ contains a collider $V_k$ such that $V_k$ is not in $\mathbf{M}$ and no descendant of $V_k$ is in $\mathbf{M}$.
	A set $\mathbf{M}$ is said to d-separate $V_i$ from $V_j$ ($ V_i \ci V_j\mid\mathbf{M}$) if and only if $\mathbf{M}$ blocks every path between $V_i$ to $V_j$. 
	Otherwise they are said to be d-connected by $\mathbf{M}$, denoted as $V_i\nindep V_j\mid\mathbf{M}$.
\end{definition}

The back-door criterion is a well-known graphical criterion for determining an adjustment set in a given DAG $\mathcal{G}$. The back-door criterion can be used directly to find an adjustment set $\mathbf{M}\subseteq \mathbf{X}$ relative to an ordered pair of variables $(V_i, V_j)$ in the given $\mathcal{G}$.

\begin{definition}[Back-door criterion~\cite{pearl2009causality}]
	\label{def:backdoorcrite}
	In a DAG $\mathcal{G}=(\mathbf{V}, \mathbf{E})$, for an ordered pair of variables $(V_i, V_j)\in \mathbf{V}$, a set of variables $\mathbf{M}\subseteq \mathbf{V}\setminus\{V_i, V_j\}$ is said to satisfy the back-door criterion in the given DAG $\mathcal{G}$ if (i) $\mathbf{M}$ does not contain a descendant node of $V_i$; and (ii) $\mathbf{M}$ blocks every back-door path between $V_i$ and $V_j$ (the paths between $V_i$ and $V_j$ starting with an arrow into $V_i$). A set $\mathbf{M}$ is referred to as a \emph{back-door set} relative to $(V_i, V_j)$ in $\mathcal{G}$ if $\mathbf{M}$ satisfies the back-door criterion relative to $(V_i, V_j)$ in $\mathcal{G}$.
\end{definition}

\subsection{Instrumental variables}
\label{subsec:IVmethods}
We follow the convention and definitions of IVs used in~\cite{hernan2006instruments,silva2017learning,cui2021semiparametric}. We assume a causal DAG $\mathcal{G}$ with the set of variables $\mathbf{V}=\mathbf{X}\cup \mathbf{U}\cup \{W, Y\}$, where $W$ is a binary treatment indicator ($w = 1$ for being treated and $w=0$ for control), $Y$ the outcome of interest, $\mathbf{X}$ a set of pretreatment variables\footnote{A variable is measured  before  applying  $W$  and observing $Y$ in a study or experiment.  Pretreatment variables can be distinguished from other variables in a real-world application by domain experts~\cite{imbens2015causal}.}, \ie $\forall X\in \mathbf{X}$, $X\notin De(W\cup Y)$ where $De(W\cup Y)$ is a shorthand of $De(W) \cup De(Y)$, and $\mathbf{U}$ the set of latent confounders. The goal of this work is to estimate the average causal effect of $W$ on $Y$ from  observational data with latent confounders.

A valid IV facilitates the identification of the causal effect of $W$ on $Y$ from data with latent confounders~\cite{angrist1995two,pearl1995testability}. A valid IV $Z$~\cite{brito2002generalized,hernan2006instruments} satisfies the three conditions as described in the Introduction section.  Given a valid IV $Z$, the causal effect of $W$ on $Y$ (referred to as $\beta_{wy}$) can be calculated as $\sigma_{zy}/\sigma_{zw}$, where $\sigma_{zy}$ and $\sigma_{zw}$ are the estimated causal effects of $Z$ on $Y$ and $Z$ on $W$, respectively. 

In this work, we employ the orthogonal instrumental variables approach (Ortho.IV)~\cite{chernozhukov2018double,syrgkanis2019machine} to calculate $\beta_{wy}$ from data with latent confounders when an IV $Z$ is available.  The Ortho.IV is to optimise the minimisation problem of a loss function that satisfies a Neyman orthogonality criterion with the aid of a known IV by solving the  moment equation~\cite{chernozhukov2018double}:
		$\mathbb{E}[(Y-\mathbb{E}[Y\mid \mathbf{X}]-\theta(\mathbf{X}) \times(W-\mathbb{E}[W\mid\mathbf{X}]))(Z-\mathbb{E}[Z\mid \mathbf{X}])] = 0$, where $\theta(.)$ is a function of $\mathbf{X}$. Once we have a valid IV, the  Ortho.IV method can be used to obtain $\beta_{wy}$ unbiasedly from observational data with latent variables. 

An IV, such as $Z$ in $Z\rightarrow W$ in Fig.~\ref{fig:exp001} (b), is often unmeasured in many real-world applications. An effect variable of an IV, such as $S$ in Fig.~\ref{fig:exp001} (c) is more likely to be measured in real-world cases, and is called a surrogate instrumental variable (SIV)~\cite{greenland2000introduction,hernan2006instruments,martens2006instrumental}.

\begin{definition}[Surrogate Instrumental Variable (SIV)]
	\label{def:SIV}
	In a causal DAG $\mathcal{G}\!\!=\!\!(\mathbf{X}\cup \mathbf{U}\cup \{W, Y\}, \mathbf{E})$, a variable $S\in\mathbf{X}$ is said to be an SIV \wrt $W\rightarrow Y$, if (i) $S$ and $W$ share a latent IV $Z$ (\ie $S\leftarrow Z\rightarrow W$), (ii) $S$ and $Y$ are associated only through $W$ (\ie exclusion restriction), and (iii) $S$ does not share common causes with $Y$ (\ie unconfoundedness instrument).
\end{definition}

An SIV is a proxy variable of a standard IV~\cite{hernan2006instruments,martens2006instrumental}. An SIV can be used as a valid IV. However, in practice, it is often difficult to identify which variable is a valid IV (standard IV or SIV) even if it is measured.

\subsection{Problem setup}
\label{subsec:proset}
We assume that the data is generated from a causal DAG $\mathcal{G}\!\!=\!\!(\mathbf{X}\cup \mathbf{U}\cup \{W, Y\}, \mathbf{E})$ containing the treatment variable $W$, the outcome variable $Y$, a set of pretreatment variables $\mathbf{X}$, and a set of latent confounders $\mathbf{U}$. There exists at least one SIV in $\mathbf{X}$, but we do not know which $X$s are SIV(s). Or the SIV information is embedded in a number of $X$s.  We will query the causal effect of $W$ on $Y$, \ie $\beta_{wy}$ from observational data. We allow the existence of multiple back-door paths between $W$ and $Y$ with some latent variables in $\mathbf{U}$ lying on some of these back-door paths between $W$ and $Y$.

\begin{figure}[t]
\centering
\includegraphics[scale=0.4]{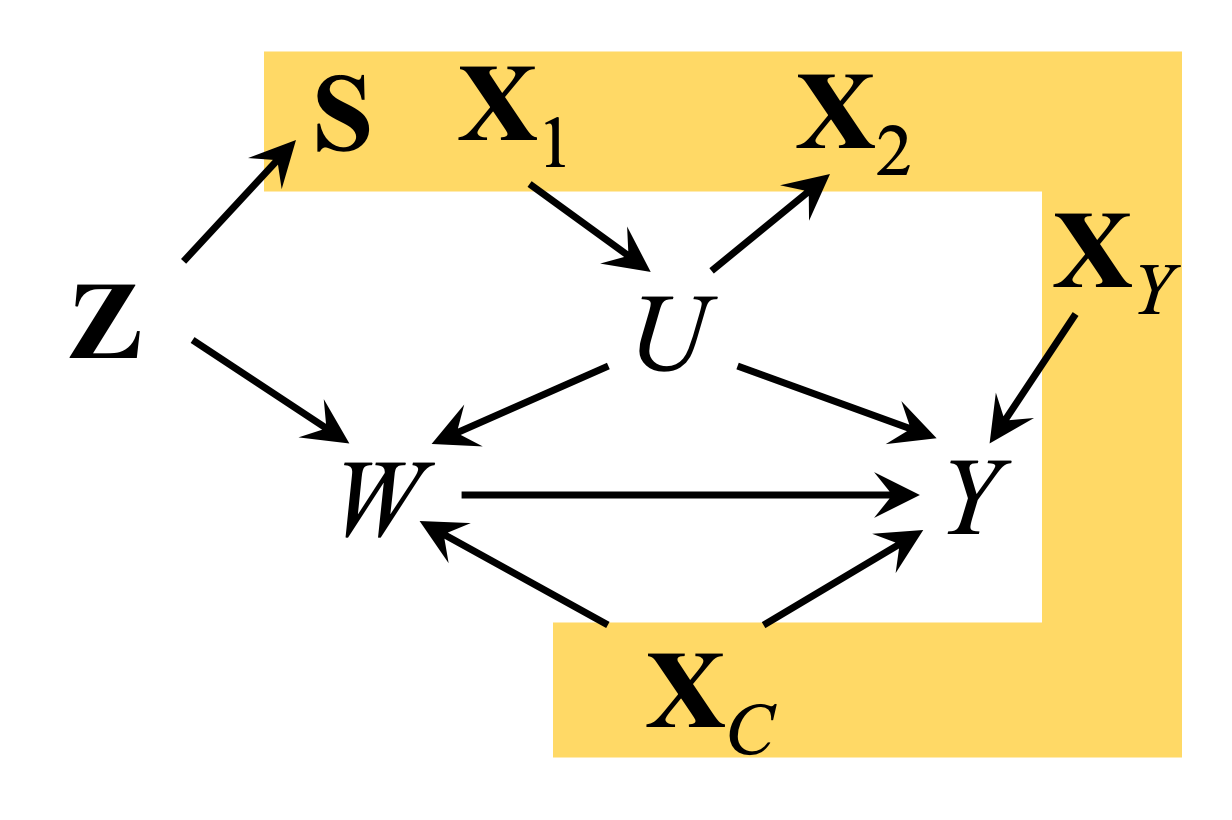}
\caption{An example causal DAG representing the data generation mechanism. The shaded area indicates all the measured pretreatment variables, and among them, $\mathbf{S}$ is a set of SIVs, $\mathbf{Z}$ is a set of latent IVs and $U$ is a latent confounder affecting both $W$ and $Y$.}
\label{fig:exp002}
\end{figure}

 We do not assume that an IV or an SIV is known, and it is difficult to identify an IV or an SIV from data. For example, assume that the causal DAG in Fig.~\ref{fig:exp002} represents a data generation mechanism. $\mathbf{S}$ is a set of SIVs. In the DAG, we can infer that $\mathbf{S}$ is independent of each variable in $\{\mathbf{X}_1, \mathbf{X}_2, \mathbf{X}_Y, \mathbf{X}_C\}$. In data, we can obtain a set of independent pairs, $(\mathbf{S}, \mathbf{X}_1), (\mathbf{S}, \mathbf{X}_2), (\mathbf{S}, \mathbf{X}_C), (\mathbf{X}_C, \mathbf{X}_1), (\mathbf{X}_C, \mathbf{X}_2), (\mathbf{X}_1,\mathbf{X}_2)$. This does not help us to find which variable is an SIV. One may wish to learn a PAG (partial ancestral graph)~\cite{richardson2002ancestral} from data for determining the set of SIVs $\mathbf{S}$. However,  two variables in the above pair (\eg $\mathbf{S}$ and $\mathbf{X}_Y$) cannot be separated since the spurious association caused by the latent confounder $U$~\cite{pearl1995testability,brito2002generalized}.  Hence, a causal discovery algorithm using conditional independence tests such as FCI (fast causal inference)~\cite{spirtes2000causation}  does not help us find an SIV from data. Even worse, the IV information might be scattered in other variables.

We will use the disentangling techniques~\cite{hassanpour2019learning,zhang2021treatment} to learn the latent IV representation through disentangling the latent representation of $\mathbf{X}$. We aim at learning a latent representation $\mathbf{\Phi}=(\mathbf{Z}, \mathbf{C})$  of $\mathbf{X}$, where $\mathbf{Z}$ represents IVs in $\mathbf{X}$, and $\mathbf{C}$ represents the remaining information in $\mathbf{X}$. Our problem setting is given below. 

\begin{problem}
Given a joint distribution $P(\mathbf{X}, W, Y)$ generated from an underlying DAG $\mathcal{G}\!\!=\!\!(\mathbf{X}\cup \mathbf{U}\cup \{W, Y\}, \mathbf{E})$. $W$ and $Y$ are treatment and outcome variables respectively. $\mathbf{X}$ is  pretreatment variables. $\mathbf{U}$ contains unobserved variables including unobserved confounders of $W$ and $Y$.   Suppose that there exists at least one SIV (\ie a set of SIVs $\mathbf{S}\subseteq\mathbf{X}$ with $|\mathbf{S}| \geq 1$). Our goal is to learn a latent IV representation $\mathbf{Z}$ through the disentanglement of the latent representation $\mathbf{\Phi}$ of $\mathbf{X}$ into two disjoint sets $(\mathbf{Z}, \mathbf{C})$ for recovering the causal effect of $W$ on $Y$.
\end{problem} 

\begin{figure}[t]
\centering
\includegraphics[scale=0.4]{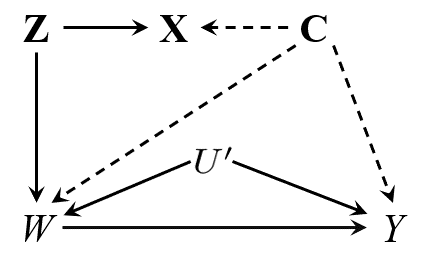}
\caption{The disentanglement scheme of DIV.VAE, represented as a causal graph. The dotted arrows indicate possible ancestral relationships between nodes. $W$, $Y$ and $U'$ are the treatment variable, the outcome and the latent confounder of $W$ and $Y$, respectively. $\mathbf{X}$ is the set of measured pretreatment variables and contains at least one SIV. $\mathbf{\Phi}=(\mathbf{Z}, \mathbf{C})$ is the latent representation of $\mathbf{X}$, where $\mathbf{Z}$ and $\mathbf{C}$ are the sets of disentangled IV representation and confounding representation, respectively.} 
\label{fig:DIV.VAE}
\end{figure}

\section{The proposed DIV.VAE method}
\label{sec:IV.DVAE}
\subsection{The proposed disentanglement scheme}
\label{subsec:graphicalproperty}
Following the literature~\cite{louizos2017causal,hassanpour2019learning,zhang2021treatment}, we propose the causal structure in Fig.~\ref{fig:DIV.VAE} to represent the causal relationships among $W$, $Y$, $U'$, $\mathbf{X}$, $\mathbf{Z}$ and $\mathbf{C}$, where the set $\mathbf{X}$ is generated from the set of latent variables $\mathbf{\Phi}=(\mathbf{Z}, \mathbf{C})$, where $\mathbf{Z}$ is the latent IV representation, and $\mathbf{C}$ captures the remaining information in $\mathbf{X}$.

We first show that the proposed disentanglement scheme is able to estimate causal effect of $W$ on $Y$ in presenting the following theorem. 	

\begin{theorem}
\label{theorem:001}
Given a joint distribution $P(\mathbf{X}, W, Y)$ generated from a causal DAG $\mathcal{G}\!\!=\!\!(\mathbf{X}\cup \mathbf{U}\cup \{W, Y\}, \mathbf{E})$. $\mathcal{G}$ contains $W\rightarrow Y$ and $W\leftarrow U'\rightarrow Y$ in $\mathcal{G}$, and $\forall X\in\mathbf{X}$, $X\notin De(W\cup Y)$ in $\mathcal{G}$. There exists at least one SIV (\ie a set of SIVs $\mathbf{S}\subseteq\mathbf{X}$ with $|\mathbf{S}| \geq 1$). If we learn and disentangle simultaneously the latent representation $\mathbf{\Phi}$ of $\mathbf{X}$ into two disjoint sets $(\mathbf{Z}, \mathbf{C})$, where $\mathbf{Z}$ is a common cause of $W$ and $\mathbf{X}$, and $\mathbf{C}$ is a common cause of  $\mathbf{X}$, $W$, and $Y$, respectively, then $\mathbf{Z}$ is a valid IV for estimating the causal effect of $W$ on $Y$ from data  of $P(\mathbf{X}, W, Y)$.
\end{theorem}
\begin{proof}
We prove that the IV representation $\mathbf{Z}$ is a valid IV based on the disentanglement causal model. First of all,  the clause $|\mathbf{S}| \geq 1$ is to ensure that there is information of valid IVs in the set of covariates $\mathbf{X}$.
In the causal DAG, (1) $\mathbf{Z}$ is a set of causes of $W$, so $\mathbf{Z}$ satisfies the first condition (i) of an IV as described in Introduction; (2) there is only a causal path from $\mathbf{Z}$ to $Y$, \ie $\mathbf{Z}\rightarrow W\rightarrow Y$, so $\mathbf{Z}$ affects $Y$ only through $W$, \ie $\mathbf{Z}$ satisfies the second condition (ii) of an IV; (3) there are four back-door paths from $\mathbf{Z}$ to $Y$, \ie $\mathbf{Z}\rightarrow \mathbf{X}\leftarrow \mathbf{C} \rightarrow Y$,  $\mathbf{Z}\rightarrow \mathbf{X}\leftarrow \mathbf{C} \rightarrow W\leftarrow U\rightarrow Y$, $\mathbf{Z}\rightarrow W\leftarrow \mathbf{C} \rightarrow Y$ and $\mathbf{Z}\rightarrow W\leftarrow U\rightarrow Y$, and all four back-door paths are blocked by $\emptyset$ according to the back-door criterion, \ie $\mathbf{Z}$ does not share common causes with $Y$, so $\mathbf{Z}$ satisfies the last condition of an IV. Therefore, $\mathbf{Z}$ is a set of valid IVs for estimating the causal effect of $W$ on $Y$ from data with latent confounders. 
\end{proof}

Theorem~\ref{theorem:001} states that the soundness of the proposed disentangled representation learning method relies on the ability to learn correct representations. The conditional clause `If we learn and disentangle simultaneously the latent representation $\mathbf{\Phi}$ of $\mathbf{X}$ into two disjoint sets $(\mathbf{Z}, \mathbf{C})$' in the theorem is an assumption that is unfortunately untestable in data. Such an assumption is used in previous VAE-based causal inference works~\cite{louizos2017causal,hassanpour2019learning,zhang2021treatment}.  Once $\mathbf{Z}$ is correctly learned, the causal effect of $W$ on $Y$ can be unbiasedly estimated from data with latent confounders. In real-world applications, $U$ may affect the learned representation of $\mathbf{C}$, but $\mathbf{C}$ is not used in the causal effect estimation and hence the uncertainty in $\mathbf{C}$ does not affect the unbiasedness of causal effect $W$ on $Y$.

 The establishment of Theorem~\ref{theorem:001} relies on the correctness of the disentanglement $\mathbf{\Phi} = (\mathbf{Z, C})$. In this work, we leverage variational autoencoders (VAEs) to optimise a variational lower bound on likelihood, enabling the learning of  $\mathbf{\Phi}$. This approach requires substantially weaker assumptions about the data generating process and the latent variable structure as in~\cite{kingma2014auto, louizos2017causal, kingma2019introduction}.  In the next section, we will introduce our proposed DIV.VAE for learning $\mathbf{\Phi}$ and disentangling $\mathbf{\Phi}$ into two disjoint sets, $(\mathbf{Z}, \mathbf{C})$.

\begin{figure*}[t]
\centering
\includegraphics[scale=0.5]{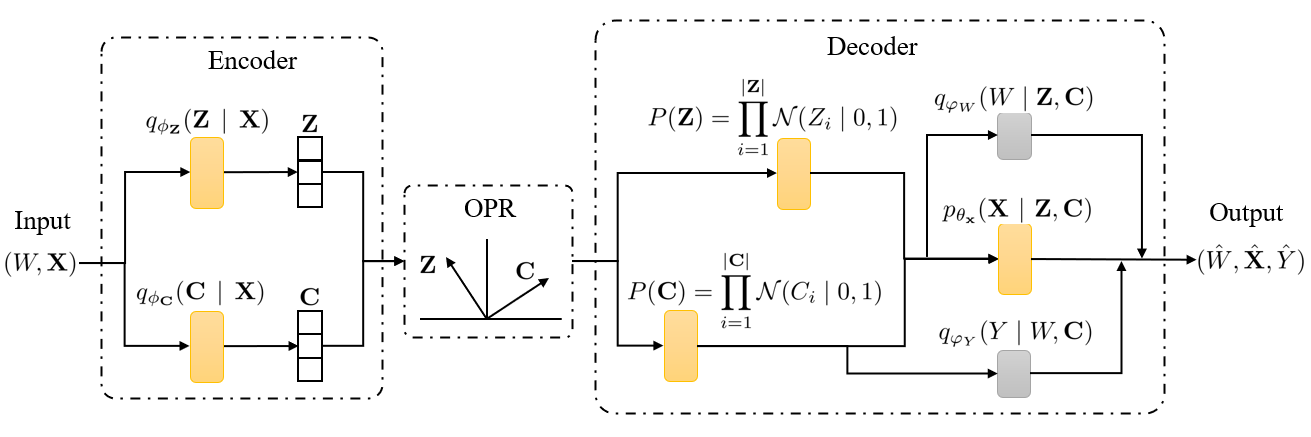}
\caption{DIV.VAE architecture. The input $\mathbf{X}$ is encoded by $q_{\phi_{\mathbf{Z}}}(\mathbf{Z}\mid\mathbf{X})$ and $q_{\phi_{\mathbf{C}}}(\mathbf{C}\mid\mathbf{X})$ into the parameters of the latent representation. The middle dashed box is the orthogonality promoting regularisation (OPR) for ensuring $\mathbf{Z}\ci\mathbf{C}$. Samples are drawn from each of the latent representations using the reparameterised trick. The samples are then concatenated and decoded through $p_{\theta_{\mathbf{x}}}(\mathbf{X}\mid\mathbf{Z}, \mathbf{C})$. The two grey boxes indicate the two auxiliary predictors $q_{\varphi_{W}}(W\mid\mathbf{Z},\mathbf{C})$ and $q_{\varphi_{Y}}(Y\mid W, \mathbf{C})$.}
\label{fig:DIVVAE_architecture}
\end{figure*}

\subsection{Finding IV representation by disentangled representation learning}
\label{subsec:model}
In this section, we introduce the details of the proposed VAE-based disentangled representation learning architecture of DIV.VAE (shown in Fig.~\ref{fig:DIVVAE_architecture}) to learn a valid IV representation $\mathbf{Z}$ following the proposed scheme in Fig.~\ref{fig:DIV.VAE}. Then we can use the learned IV representation to obtain unbiased causal effect estimation from data with latent confounders. 

The goal of our designed architecture for DIV.VAE is to learn  the latent representation $\mathbf{\Phi}$ of $\mathbf{X}$ and disentangle $\mathbf{\Phi}$ into $(\mathbf{Z, C})$ simultaneously, following the proposed causal structure in Fig.~\ref{fig:DIV.VAE}. It is worth noting that $\mathbf{C}$ plays a critical role as a set of auxiliary variables used to capture the information from the set of $\mathbf{X}\setminus\{\mathbf{S}\}$  in the representation learning and disentanglement process but it is not used for the causal effect estimation.  

The proposed DIV.VAE architecture in Fig.~\ref{fig:DIVVAE_architecture} uses the inference model and the generation model to approximate the posterior $p_{\theta_{\mathbf{X}}}(\mathbf{X}\mid\mathbf{Z},\mathbf{C})$ where $\theta$ is a set of generative model parameters. For the inference model, we develop two separate encoders $q_{\phi_\mathbf{Z}}(\mathbf{Z}\mid\mathbf{X})$ and $q_{\phi_\mathbf{C}}(\mathbf{C}\mid\mathbf{X})$ that serve as variational posteriors over the latent variables. For the generative model, the two latent representations $(\mathbf{Z},\mathbf{C})$ are obtained from the two separate encoders used by a single decoder $p_{\theta_{\mathbf{X}}}(\mathbf{X}\mid\mathbf{Z},\mathbf{C})$ to reconstruct $\mathbf{X}$. Following the VAE literature~\cite{kingma2014auto,kingma2019introduction}, the prior distributions of $P(\mathbf{Z})$ and $P(\mathbf{C})$ are drawn from Gaussian distributions.

In the inference model, the variational approximations of the posteriors are defined as:
\begin{equation}
\label{eq:001}
\begin{aligned}
	&q_{\phi_{\mathbf{Z}}}(\mathbf{Z}\mid\mathbf{X}) = \prod_{i=1}^{\left|\mathbf{Z} \right|} \mathcal{N}(\mu = \hat{\mu}_{Z_i}, \sigma^2 = \hat{\sigma}^2_{Z_i}); \\& q_{\phi_{\mathbf{C}}}(\mathbf{C}\mid\mathbf{X}) = \prod_{i=1}^{\left|\mathbf{C}\right|} \mathcal{N}(\mu = \hat{\mu}_{C_i}, \sigma^2 = \hat{\sigma}^2_{C_i}),
\end{aligned}	
\end{equation}
\noindent where $\hat{\mu}_{Z_i},\hat{\mu}_{C_i}$ and $\hat{\sigma}^2_{Z_i},\hat{\sigma}^2_{C_i}$ are the means and variances of the Gaussian distributions parameterised by neural networks. Note that, since one IV is sufficient for obtaining unbiased causal effect estimation, we use $\left|\mathbf{Z}\right| = 1$ in the experiments on real-world datasets. However, in the algorithm design, we keep $\mathbf{Z}$ as multi-dimensional for a general solution.

The prior distributions of $(\mathbf{Z, C})$ are defined as:
\begin{equation}
\label{eq:002}
P(\mathbf{Z}) = \prod_{i=1}^{\left|\mathbf{Z} \right|} \mathcal{N}(Z_{i} \mid 0,1);  P(\mathbf{C}) = \prod_{i=1}^{\left|\mathbf{C} \right|} \mathcal{N}(C_{i} \mid 0,1).
\end{equation} 

The generative models for $W$ and $\mathbf{X}$ are defined as:
\begin{equation}
\label{eq:003}
\begin{aligned}
	&p_{\theta_{W}}(W\mid\mathbf{Z}, \mathbf{C}) = Bern(\sigma(g_1(\mathbf{Z}, \mathbf{C})));  \\& p_{\theta_{\mathbf{x}}}(\mathbf{X}\mid\mathbf{Z}, \mathbf{C}) = \prod_{i=1}^{\left|\mathbf{X} \right|} P(X_i \mid\mathbf{Z}, \mathbf{C}),
\end{aligned}	
\end{equation} 
\noindent where $P(X_i \mid\mathbf{Z}, \mathbf{C})$ is the distribution for the $i$th measured variable, $g_1(\cdot)$ is the function parameterised by neural networks and $\sigma(\cdot)$ is the logistic function.

The generative model for $Y$ depends on the data type of $Y$. For continuous $Y$, we sample it from a Gaussian distribution with its mean and variance given by the mutually exclusive neural networks that define $P(Y \mid W = 0, \mathbf{Z}, \mathbf{C})$ and $P(Y \mid W = 1, \mathbf{Z}, \mathbf{C})$ respectively, and the generative model of $Y$ is defined as:
\begin{equation}
\label{eq:004}
\begin{aligned}
	& P(Y \mid W, \mathbf{C}) = \mathcal{N}(\mu = \hat{\mu}_{Y}, \sigma^2 = \hat{\sigma}^2_{Y}); \\
	&\hat{\mu}_{Y} = W \cdot g_2(\mathbf{C}) + (1-W) \cdot g_3(\mathbf{C});  \\ 
	&\hat{\sigma}^2_{Y} = W \cdot g_4(\mathbf{C}) + (1-W) \cdot g_5(\mathbf{C}),
\end{aligned}	
\end{equation}
\noindent where $g_2(\cdot)$, $g_3(\cdot)$, $g_4(\cdot)$ and $g_5(\cdot)$ are the functions parameterised by neural networks. For binary $Y$, we parameterise it with a Bernoulli distribution and the model is defined as:
\begin{equation}
\label{eq:005}
\begin{aligned}
	&p_{\theta_{Y}}(Y\mid W, \mathbf{C}) = Bern(\sigma(g_6(W, \mathbf{C}))),
\end{aligned}	
\end{equation}
\noindent where $g_6(\cdot)$ is a neural network with its own parameters. Given the joint distribution $P(\mathbf{X}, W, Y)$, the parameters can be optimised by maximising the evidence lower bound (ELBO) $\mathcal{M}$~\cite{kingma2014auto}:
\begin{equation}
\label{eq:006}
\begin{aligned}
	\mathcal{M} = \: &\mathbb{E}_{q_{\phi_{\mathbf{Z}}} q_{\phi_{\mathbf{C}}}}[\log p_{\theta_{\mathbf{x}}}(\mathbf{X}\mid\mathbf{Z}, \mathbf{C})] \\& -   D_{KL}[q_{\phi_{\mathbf{Z}}}(\mathbf{Z}\mid \mathbf{X})||P(\mathbf{Z})]  \\ 
	& - D_{KL}[q_{\phi_{\mathbf{C}}}(\mathbf{C}\mid\mathbf{X})||P(\mathbf{C})],
\end{aligned}	
\end{equation}
\noindent where $D_{KL}[\cdot||\cdot]$ is a KL divergence term. 

To learn the latent IV representation $\mathbf{Z}$ from the set of SIVs $\mathbf{S}$ and the latent representation $\mathbf{C}$ from the remaining variables $\mathbf{X}\setminus\{\mathbf{S}\}$, we add two auxiliary predictors to the above variational ELBO to ensure that the treatment variable $W$ and the outcome variable $Y$ can be estimated from $\mathbf{Z}$ and $\mathbf{C}$ as designed. Thus, we have the following objective function:
\begin{equation}
\label{eq:007}
\begin{aligned}
	\mathcal{L}' =\: & -\mathcal{M} + \alpha_W \mathbb{E}_{q_{\phi_{\mathbf{Z}}}q_{\phi_{\mathbf{C}}}}[\log q_{\varphi_{W}}(W\mid\mathbf{Z}, \mathbf{C})]\\&~~ + 
	\alpha_Y \mathbb{E}_{q_{\phi_{\mathbf{C}}}}[\log q_{\varphi_{Y}}(Y\mid W, \mathbf{C})],
\end{aligned}	
\end{equation}
\noindent where $\alpha_{W}$ and $\alpha_{Y}$ are the weights for the two auxiliary predictors. 

In practice, there may be some very weak associations between $\mathbf{Z}$ and $\mathbf{C}$. To encourage $\mathbf{Z}\ci\mathbf{C}$ as specified in Fig.~\ref{fig:DIV.VAE}, we employ the orthogonality promoting regularisation (OPR)~\cite{xie2018orthogonality} for our proposed DIV.VAE:	
\begin{equation}
\label{eq:008}
\begin{aligned}
	\mathcal{L} =\: & \mathcal{L}' + \frac{1}{b}\sum_{i=1}^{b}CS(\mathbf{Z}_i,\mathbf{C}_i),
\end{aligned}	
\end{equation}
\noindent where $b$ is the batch size of the neural network, and $CS(\mathbf{Z}_i, \mathbf{C}_i)= \frac{\mathbf{Z}_i^{T}\mathbf{C}_i}{\left\|\mathbf{Z}_i\right\|_2 \left\|\mathbf{C}_i\right\|_2}$ is the cosine similarity (CS). 

After training DIV.VAE, we draw $\mathbf{Z}$ from the model and feed it into the function of Ortho.IV~\cite{chernozhukov2018double,syrgkanis2019machine} for calculating $\beta_{wy}$. When the learned distribution of $\mathbf{Z}$ is close to the true unmeasured IV distribution, the DIV.VAE method has the ability to obtain an unbiased estimate $\beta_{wy}$ as shown in the experimental results. Notably, the main advantage of our DIV.VAE is that it no longer requires domain knowledge or experts to provide a valid IV. Instead, it only requires the presence of a SIV in the data. This is a weaker assumption compared to those required by other methods such as TSLS, FIVR, DeepIV, and IV.Tetrad. 

\noindent\textbf{Limitations}. The soundness of DIV.VAE relies on the ability of the proposed VAE architecture to learn $\mathbf{\Phi}$ and disentangle the latent variable $\mathbf{\Phi}$ into $(\mathbf{Z, C})$. However, VAE-based methods are susceptible to the problem of unidentifiability in the VAE model~\cite{locatello2019challenging, khemakhem2020variational}. In other words, there is no theoretical guarantee that the learned IV representation $\mathbf{Z}$ can always approximate the true latent IV. Fortunately, as shown in our experiments, in the presence of SIV, the learned IV representation $\mathbf{Z}$ by DIV.VAE closely approximates the true latent IV.  We note that iVAE gives an identifiability guarantee but with more limitations~\cite{khemakhem2020variational}. iVAE assumes injective and linear relationships and a latent presentation learned by iVAE needs its child variable and parent variables to be observed. These additional requirements limit the application of a method based on iVAE and make it infeasible for iVAE to recover the IV representation from the error term of SIV.  VAE does not guarantee the identifiability but has some advantages, such as not requiring linear and injective relationships or $\mathbf{Z}$'s parent to be observed for its representation learning. However, users should review their results by DIV.VAE with domain knowledge and perform sensitivity analyses~\cite{imbens2015causal} before taking the results.

\section{Experiments}
\label{sec:exp}
\subsection{Experiment setup}
\label{subsec:expset}
\textbf{Baseline causal effect estimators}. We compare DIV.VAE with four representative IV based estimators and two VAE-based causal effect estimators. Three of the IV based estimators, two-stage least squares (TSLS) regression~\cite{angrist1995two}, causal random forest for IV regression (FIVR)~\cite{athey2019generalized}, and the popular deepIV~\cite{hartford2017deep}, each requires a given IV; whereas the other IV based estimator, IV.Tetrad~\cite{silva2017learning} does not require a given IV, but needs the majority of variables in $\mathbf{X}$ to be valid IVs. The two VAE-based estimators are causal effect variational autoencoder (CEVAE)~\cite{louizos2017causal} and treatment effect by disentangled variational autoencoder (TEDVAE)~\cite{zhang2021treatment}. These two estimators assume no latent confounder between $(W, Y)$. They have been used in our experiments since our DIV.VAE is also based on the VAE model.

\textbf{Evaluation metrics}. For synthetic datasets with the ground-truth $\beta_{wy}$, we use the estimation bias $\left |(\hat{\beta}_{wy}-\beta_{wy})/\beta_{wy} \right|*100$ (\%) to demonstrate the performance of all estimators. For the real-world datasets, we evaluate all estimators against the reference causal effects in the literature. 

\textbf{Implementation details}. DIV.VAE is implemented by using Python with packages including \textit{pytorch}~\cite{paszke2019pytorch}, \textit{pyro}~\cite{bingham2019pyro} and \textit{econML}~\cite{battocchi2019econml}. 
The implementation of TSLS is based on the functions \textit{glm} and \textit{ivglm} in the R packages \textit{stats} and \textit{ivtools}~\cite{sjolander2019instrumental}. FIVR is implemented using the function \textit{instrumental forest}  in the R package \textit{grf}~\cite{athey2019generalized}. DeepIV is retrieved from the authors' GitHub~\footnote{\url{https://github.com/jhartford/DeepIV}}. IV.Tetrad is also retrieved from the authors' site~\footnote{\url{http://www.homepages.ucl.ac.uk/~ucgtrbd/code/iv_discovery}}. CEVAE is implemented using the function \textit{CEVAE} in the Python package \textit{pyro}~\cite{bingham2019pyro} and TEDVAE is obtained from the authors' GitHub~\footnote{\url{https://github.com/WeijiaZhang24/TEDVAE}}.

In our experiments, the performance of DIV.VAE against the above baselines on simulated and real-world data. We will provide the ablation experiments of our DIV.VAE and the empirical evaluation on the independence relation of $\mathbf{Z}$ and $\mathbf{C}$ in Sections~\ref{subsec:as} and~\ref{subsec:zc}, respectively.

\subsection{Simulation study}
\label{subsec:simulationstudy}
\begin{figure}[t]
	\centering
	\includegraphics[scale=0.35]{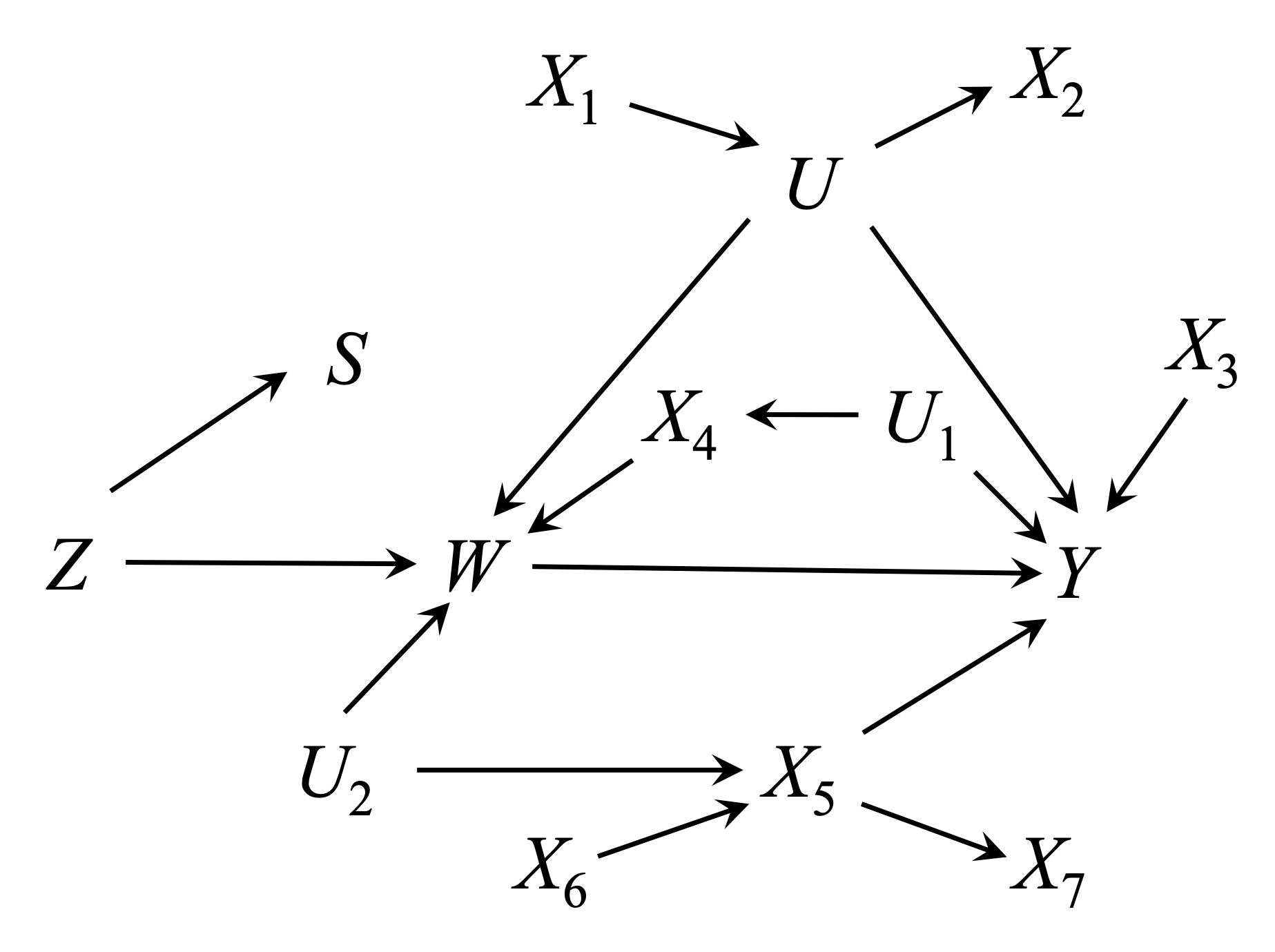}
	\caption{The true causal DAG with a latent confounder $U$ between $W$ and $Y$ is used to generate the synthetic datasets. $Z$ and $S$ are a latent IV and an SIV, respectively. $\{U_1, U_2\}$ are two latent variables, and other measured variables are pretreatment variables of $(W, Y)$.}
	\label{fig:figure_001}
\end{figure}

We utilise the true DAG over $\mathbf{X}\cup\mathbf{U}\cup\{W, Y\}$ as shown in Fig.~\ref{fig:figure_001} to generate the synthetic datasets with latent variables for the experiments by following the literature~\cite{cheng2022ancestral}. We generate datasets with a range of sample sizes:  0.5k, 2k, 4k, 6k, 8k,  10k, 50k, 100k and 200k. The set of measured variables $\mathbf{X}$ consists of $\{S, X_1, X_2, X_3, X_4, X_5, X_6, X_7, W, Y\}$. The DAG also include three latent variables $\{U, U_1, U_2\}$ in the data, where $U$ affects both $W$ and $Y$. The details of data generation are introduced as follows.	

 The synthetic datasets are generated based on the DAG in Fig.~\ref{fig:figure_001}, and the specifications are as follows:
	$Z \sim N(0, 1);~~U_1, U_2 \sim N(0, 1);
	X_1, X_3, X_5, X_7 \sim N(0, 1);
	\epsilon_{1, 2, 3, 4, S} \sim N(0, 0.5);
	S\sim N(0, 1) + Z+\epsilon_{S};
	U\sim N(0, 1) + 0.8*X_1+\epsilon_{1}; 
	X_2 \sim N(0, 1) +2*U+\epsilon_{2};
	X_4\sim N(0, 1)+ U_1 +\epsilon_{3};
	X_6 \sim N(0, 1) +0.6* U_2 +\epsilon_{4}$, where $N( , )$ denotes the normal distribution. The treatment assignment $W$ is generated from $n$ ($n$ is the sample size) Bernoulli trials by using the assignment probability  based on the measured variables $\{X_4, X_5\}$ and latent variables $\{U, U_2\}$ as $P(W=1\mid U, Z, X_4, X_5, U_2)= [1+exp\{2- 2*U- 2*Z- 3*X_4- X_5 -3*U_2\}]^{-1}$.

\begin{figure*}[t]
	\centering
	\includegraphics[scale=0.45]{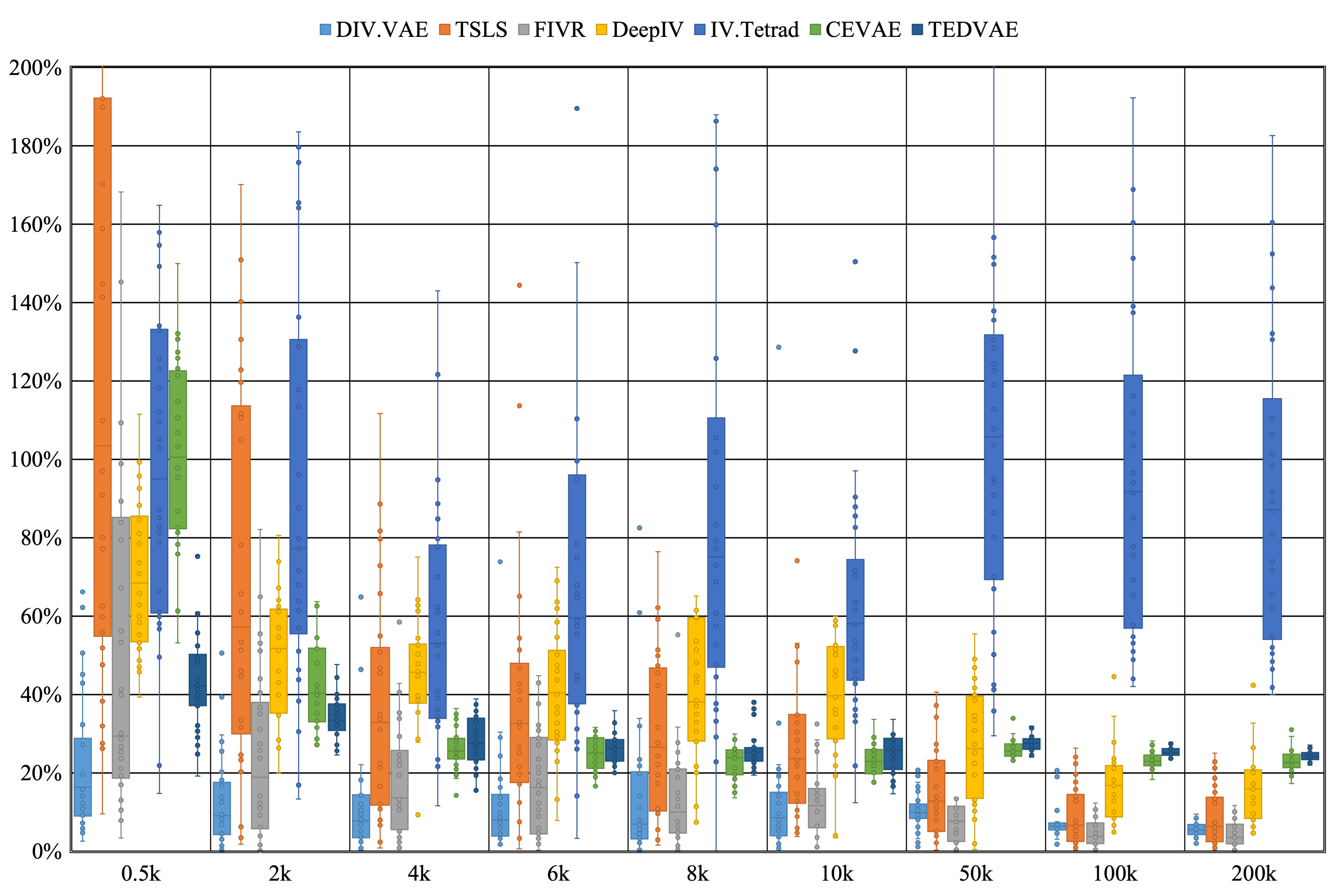}
	\caption{Estimation biases over 30 synthetic datasets with $Y_{linear}$ for different estimators, where the horizontal axis represents the sample size and the vertical axis represents the estimation bias (\%). FIVR performs competitively with DIV.VAE in large datasets. FIVR needs a given IV whereas DIV.VAE does not.} 
	\label{fig:syntheticresults}
\end{figure*}

\begin{figure*}[t]
	\centering
	\includegraphics[scale=0.45]{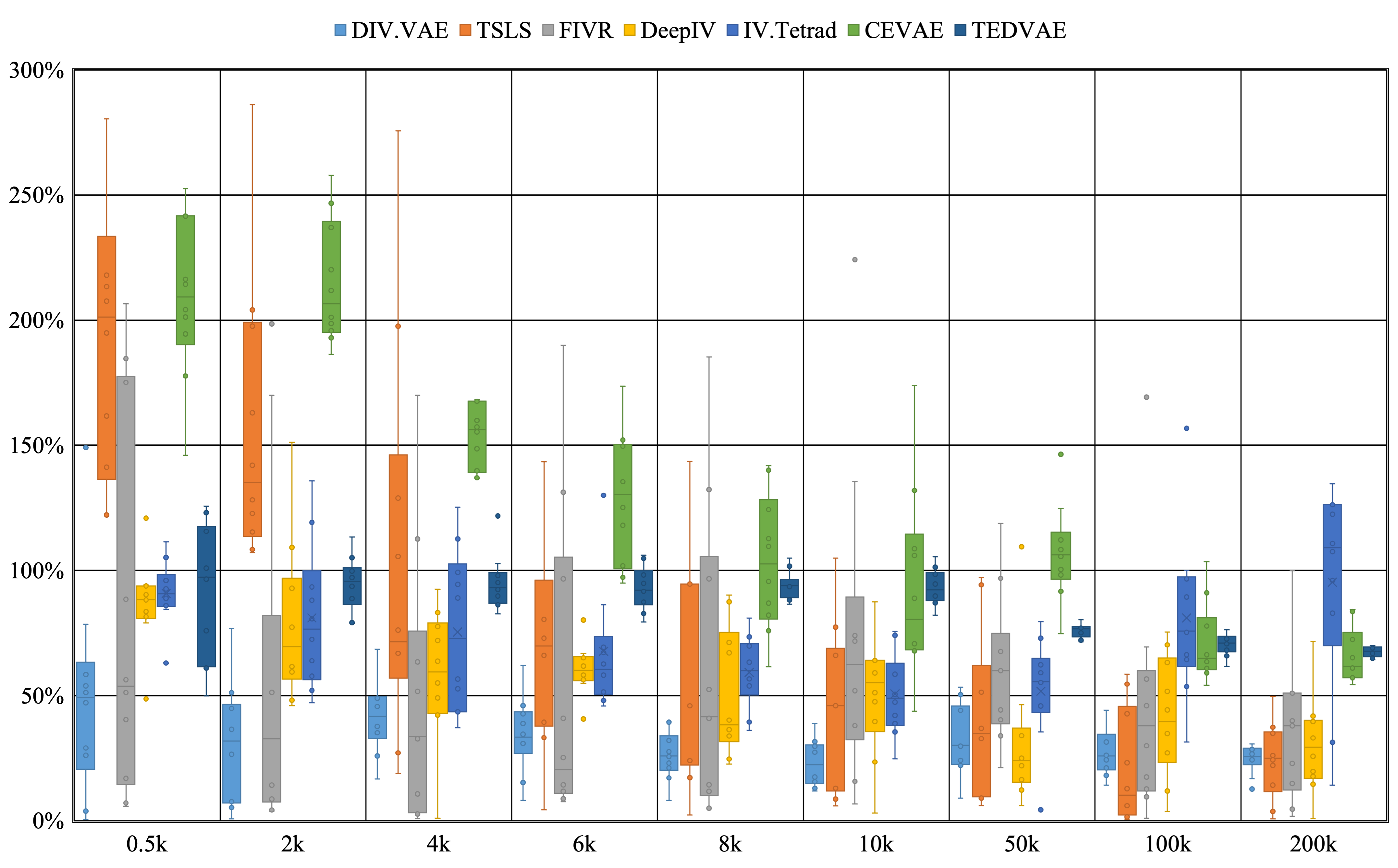}
	\caption{Estimation biases  over 30 synthetic datasets with $Y_{nonlinear}$ for different estimators, where the horizontal axis represents the sample size and the vertical axis represents the estimation bias (\%). DeepIV performs competitively with DIV.VAE in large datasets. DeepIV needs a given IV whereas DIV.VAE does not.}
	\label{fig:syntheticresults_nonlinear}
\end{figure*}

 In this work, we generate two types of potential outcomes, i.e.  a linear function $Y_{linear}$ and a non-linear function $Y_{nonlinear}$ for evaluating the ability of DIV.VAE in terms of causal effect estimation. $Y_{linear}= 2 + 2*W + 2*U + 3*U_1 + 2*X_3 + 2*X_6  + 2*X_7+\epsilon_{w}$,  where $\epsilon_{w}\sim N(0,1)$, and  $Y_{nonlinear} = 2 + 2*W + 2*U + 3*U_1 + 2*X_3 + 2*X_6^{2}  + 2*X_7+\epsilon_{w}$. Note that the causal effect of $W$ on $Y$ is fixed to 2, \ie $\beta_{wy}=2$ on all synthetic datasets.

To avoid the random noises brought by data generation process, we repeatedly generate 30 datasets for each sample size. In our simulation experiments, we use the SIV $S$ in the underlying causal DAG as the known IV for the compared IV based estimators, TSLS, FIVR and DeepIV.

\paragraph{Performance of DIV.VAE on causal effect estiamtion} The estimation biases of all estimators on  synthetic datasets with $Y_{linear}$ and synthetic datasets $Y_{nonlinear}$ are visualised with boxplots in  Fig.~\ref{fig:syntheticresults} and Fig.~\ref{fig:syntheticresults_nonlinear}, respectively.

\textbf{Results}. From the experimental results, we have the following observations:  (1) On all synthetic datasets, DIV.VAE consistently exhibits low bias and small variance, outperforming all compared estimators as the sample size increases. (2) DIV.VAE and FIVR both have low bias across all datasets with $Y_{linear}$, but the performance of DIV.VAE is more stable than FIVR. For smaller sized datasets, DIV.VAE has a smaller variance and bias than FIVR. This is because FIVR uses SIV which is a proxy of IV and this results in large variance with finite samples. DIV.VAE uses $\mathbf{Z}$, the representation of the IV. When the representation is learned properly, the estimation of DIV.VAE is unbiased.  (3) For the two VAE-based estimators, CEVAE and TEDVAE, they have relatively low variances, but compared to DIV.VAE, their biases are much larger on both types of synthetic datasets since both methods do not allow a latent confounder $U$ between $(W, Y)$. (4) TSLS and DeepIV  exhibit small biases on both types of datasets since they use the true SIV as a valid IV. (5) IV.Tetrad performs consistently poorly since the majority of valid IV assumption does not hold on both types of datasets.

\begin{figure}[t]
	\centering
	\includegraphics[scale=0.35]{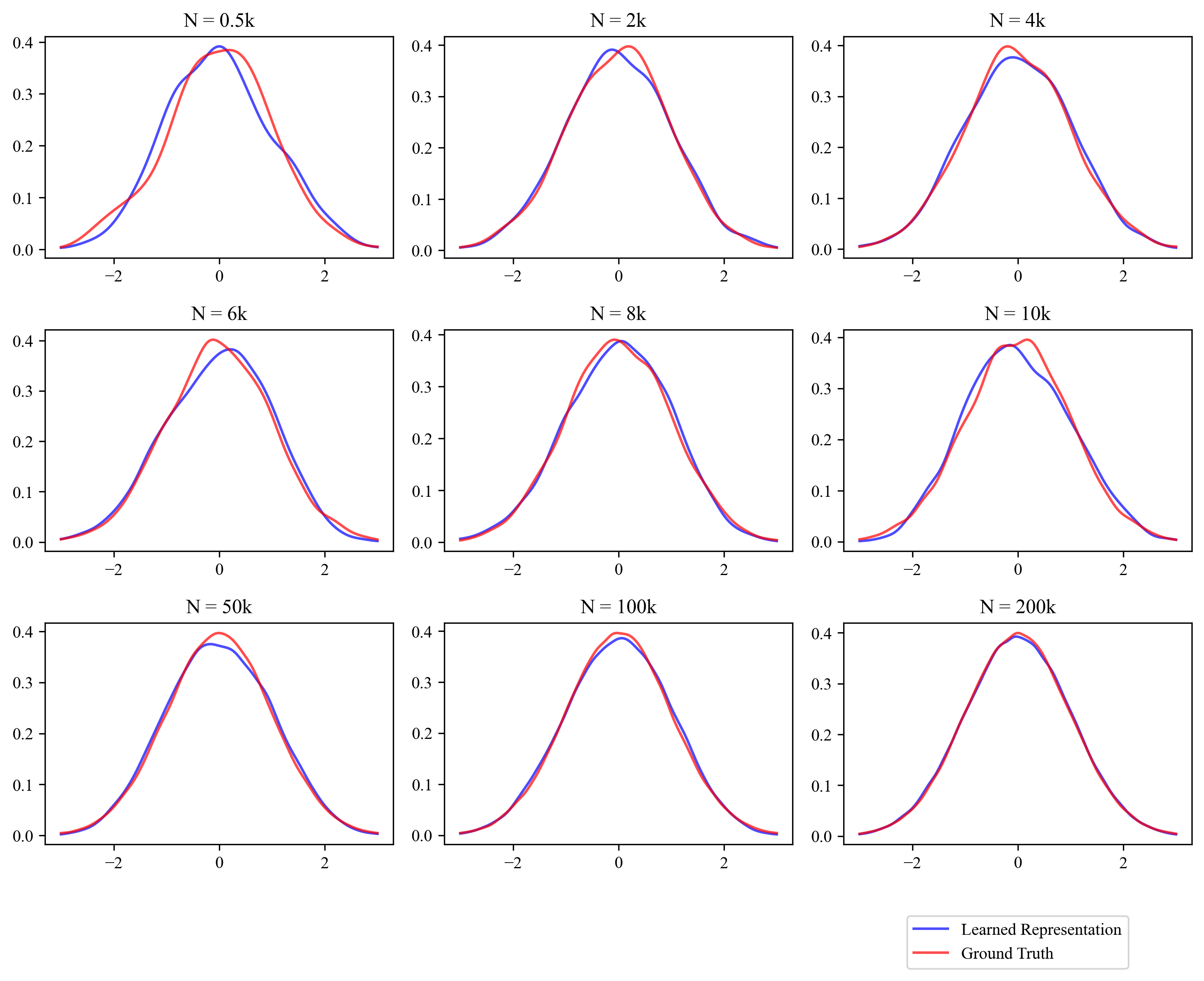}
	\caption{The probability density functions (PDFs) of the ground truth IV and the learned representation $Z$, where the horizontal axis represents the value and the vertical axis represents the density. } 
	\label{fig:correctnesslinear}
\end{figure}

\paragraph{Correctness of learned IV representation  $\mathbf{Z}$}
We examine the quality of the learned representation $\mathbf{Z}$ by visualising the  probability density functions (PDFs) of the ground truth IV and the learned IV representation $\mathbf{Z}$. We use the learned IV representation $\mathbf{Z}$ from the data with $Y_{linear}$ for the visualisation in Fig.~\ref{fig:correctnesslinear}. We have the following observations from Fig.~\ref{fig:correctnesslinear}:  (1) the learned IV representation $\mathbf{Z}$ approximates the ground truth PDF very well even when the sample size is small. (2) As the sample size increases,  the PDF of learned IV representation $\mathbf{Z}$ closely matches the  ground truth PDF. Thus, DIV.VAE is able to learn the correct IV representation from data with latent confounders.

To sum up, DIV.VAE is effective in achieving accurate and stable causal effect estimation from data without giving an IV.

\subsection{Experiments on three real-world datasets}
\label{Subsec:realworlddatasets}
In this section, we conduct experiments on two commonly used benchmark datasets with known IVs, Vitamin D (VitD)~\cite{martinussen2019instrumental}  and Schooling Returns~\cite{card1993using}. The empirical estimates of the causal effects of the two datasets are widely accepted. Another  dataset, Sachs is from a real application~\cite{sachs2005causal}. There is no nominated IV variable for Sachs.

\begin{table*}[ht]
	\centering
	\caption{Results of all estimators on the three real-world datasets. The estimated causal effects within the 95\% confidence interval are highlighted. `-' indicates that a method is not applicable since no IV is given on the Sachs dataset.}
	\begin{tabular}{ccccccccc}
		\toprule
		Estimators       & & DIV.VAE & TSLS  & FIVR  & DeepIV & IV.Tetrad & CEVAE & TEDVAE \\ \cmidrule{1-1} \cmidrule{3-9} 
		VitD             & & \textbf{3.543}   & \textbf{3.940} & 0.205 & 0.012  & \textbf{2.150}     & 0.006 & -0.042 \\ \cmidrule{1-1} \cmidrule{3-9} 
		Schoolingreturns & & \textbf{0.175}   & 0.504 & 1.151 & -0.027 & \textbf{0.064}     & \textbf{0.068} & -0.021 \\ \cmidrule{1-1} \cmidrule{3-9} 
		Sachs            & & \textbf{1.303}   & -     & -     & -      & \textbf{1.430}     & {0.239} & {0.254}  \\ \bottomrule
	\end{tabular}
	\label{tab:results}
\end{table*}

\textbf{Vitamin D data}. VitD is a cohort study of vitamin D on mortality reported in~\cite{martinussen2019instrumental}. The data contains 2,571 individuals and 5 variables: age, filaggrin (a binary variable indicating filaggrin mutations), vitd (a continuous variable measured as serum 25-OH-D (nmol/L)), time (follow-up time), and death (binary outcome indicating whether an individual died during follow-up)~\cite{sjolander2019instrumental}. The measured value of vitamin D less than 30 nmol/L implies vitamin D deficiency. The indicator of filaggrin is used as an instrument~\cite{martinussen2019instrumental}. We take the estimated $\hat{\beta}_{wy} = 2.01$ with 95\% conditional interval (0.96, 4.26) from the work~\cite{martinussen2019instrumental} as the reference causal effect.

\textbf{Schoolreturning}. The data is from the national longitudinal survey of youth (NLSY), a well-known dataset of US young employees, aged range from 24 to 34~\cite{card1993using}. The treatment is the education of employees, and the outcome is raw wages in 1976 (in cents per hour). The data contains 3,010 individuals and 19 covariates. The covariates include experience (Years of labour market experience), ethnicity (Factor indicating ethnicity), resident information of an individual, age, nearcollege (whether an individual grew up near a 4-year college?), marital status, Father's educational attainment, Mother's educational attainment, and so on. A goal of the studies on this dataset is to investigate the causal effect of education on earnings. Card~\cite{card1993using} used geographical proximity to a college, \ie the covariate \emph{nearcollege} as an instrument variable. We take $\hat{\beta}_{wy} = 0.1329$ with 95\% conditional interval (0.0484, 0.2175) from~\cite{verbeek2008guide} as the reference causal effect.

\textbf{Sachs}. This data is collected from cell activity measurements for single cell data under a variety of conditions~\cite{sachs2005causal}.  Following the work~\cite{silva2017learning}, we focus on a single condition, \ie simulation with anti-CD3 and anti-CD28. The data contains 853 records and 11 variables~\cite{sachs2005causal}. The treatment is the manipulation of concentration levels of molecule $Erk$. The outcome is the concentration of $Akt$. The other 9 cell products are pretreatment variables~\cite{sachs2005causal}. The data has some weak correlations among variables, but we assume that there are no conditional independencies held between  $Erk$ and the remaining 10 variables. Note that there is not a given instrumental variable. We take the estimated $\hat{\beta}_{wy} =1.43$ from the literature~\cite{silva2017learning} (\ie IV.Tetrad's estimated value) as the reference causal effect.

\textbf{Results}. From the results in Table~\ref{tab:results}, we see the estimated causal effects of DIV.VAE for VitD and Schoolingreturns are in their empirical intervals. On Sachs, the estimated causal effects by DIV.VAE is close to IV.Tetrad's estimated value. These results confirm that DIV.VAE is capable of recovering a latent IV representation from data. The causal effects estimated by IV.Tetrad are in the empirical intervals of VitD and Schoolingreturns since both datasets are low-dimensional and satisfy the assumptions of IV.Tetrad. The other baselines either work well on VitD, or work well on Schoolingreturns, but not on both. The Sachs dataset is not applicable to TSLS, FIVR and DeepIV since the dataset does not have a nominated IV. The two VAE-based methods do not work well on the Sachs dataset.

In sum, DIV.VAE, without needing a nominated IV, performs better or competitively with the state-of-the-art IV based or VAE-based estimators on the three real-world datasets, further confirming the effectiveness of the proposed DIV.VAE method and suggesting the potential of DIV.VAE in real-world applications.

\subsection{Evaluation in Higher Dimension with Tabular Data}
\label{subsec:highdimens}
 To evaluate the performance of our DIV.VAE with higher dimensional datasets, we generate synthetic datasets with a range of sample sizes: 0.5k, 2k, 4k, 6k, 8k, and 10k, and varying the number of measured variables as 8, 16, 32, and 64 using the same process described in Section~\ref{subsec:simulationstudy}.

Note that 64 variables are not considered as high dimensional in general machine learning settings. In causal effect estimation, however, the variables are not many since pretreatment variables are handpicked by domain experts~\cite{imbens2015causal}. We do not run DIV.VAE in a high-dimensional setting due to the faithfulness assumption it requires. Higher dimensionality is not a problem for DIV.VAE, but will pose a problem for simulation data generation. A dataset generated needs to be faithful to the underlying DAG for data generation, and to preserve the conditional independencies among a large number of variables, the size of the dataset (i.e. the number of samples) needs to be very large. A dataset with a large number of samples takes a long time for representation learning with VAE. This is why we just vary the number of variables upto 64.   
	
We use the synthetic datasets with $Y_{linear}$ to examine the performance of DIV.VAE in this section.  For each setting, we generate 30 datasets repeatedly to reduce the impact of random noise, as described in Section~\ref{subsec:simulationstudy}.

\begin{table*}[t]
	\centering
	\caption{Estimation bias of DIV.VAE in each setting over 30 synthetic datasets (mean± std).}
	\begin{tabular}{ccccccc}
				\toprule
		\multirow{2}{*}{Dimensions of variables} & \multicolumn{6}{c}{Sample sizes}                                                                    \\  \cline{2-7} 
		& 0.5k             & 2k            & 4k            & 6k            & 8k           & 10k         \\ \midrule
		8                           & 18.15 ± 39.36  & 17.25 ± 21.25 & 14.12 ± 19.11 & 16.83 ± 11.66 & 15.7 ± 24.66 & 13.88 ± 7.65 \\
		16                        & 23.74 ± 10.07 & 27.57 ± 4.2   & 25.03 ± 5.52  & 22.54 ± 5.71  & 20.48 ± 4.22  & 16.55 ± 3.42          \\
		32                    & 31.65 ± 18.72 & 30.39 ± 6.65  & 27.88 ± 5.21  & 25.04 ± 9.56  & 21.44 ± 5.38  & 18.16 ± 8.22         \\
		64                      & 36.64 ± 12.7  & 30.63 ± 14.14 & 31.14 ± 12.19 & 26.56 ± 13.12 & 23.29 ± 13.73 & 18.56 ± 11.64  \\ \bottomrule
	\end{tabular}
		\label{tab:highdimensions}
\end{table*}

\textbf{Results.}  We report the experimental results in Table~\ref{tab:highdimensions}. We have the following observations: (1) As the number of variable increases, DIV.VAE has a large estimation bias. This is because a higher dimensional dataset needs a significantly larger dataset to ensure that the dataset and the underlying causal DAG are faithful to each other. When the faithfulness assumption is not satisfied, both bias and variance of estimates are large.  (2) An increase in the number of samples results in a decrease in the bias. The reason is the same as before.  Therefore, in the case of handling high-dimensional data, DIV.VAE requires a large sample size to reduce estimation bias.

\subsection{The Capability of DIV.VAE in Image Data}
\label{subsec:capcity}
\begin{figure}[t]
\centering
\includegraphics[scale=0.32]{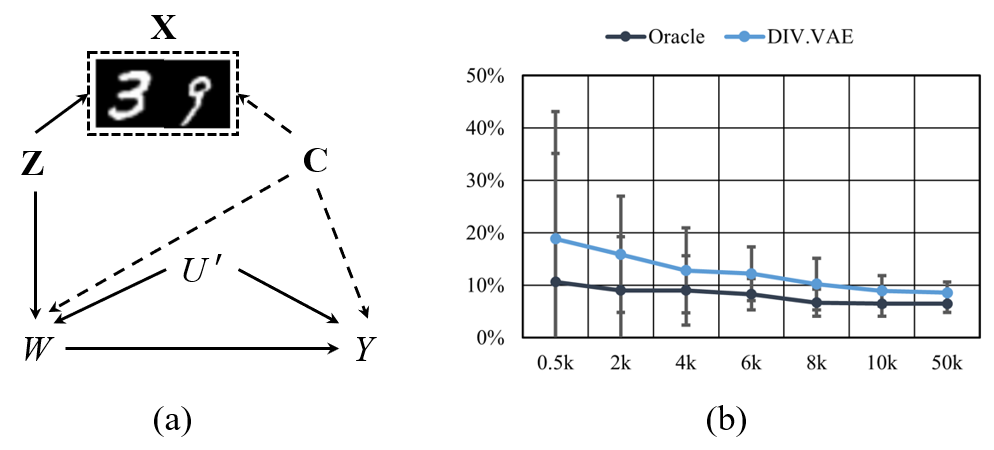}
\caption{(a) The true causal DAG for image data. $\mathbf{X}$ is replaced by image data, i.e., an image of two handwritten digits. (b) Estimation biases over 30 times for image data, where the horizontal axis represents the sample size and the vertical axis represents the estimation bias (\%).}
\label{fig:figure_image}
\end{figure}

In the previous sections, we demonstrated the performance of DIV.VAE on both simulated and real-world relational datasets. However, in many applications, we do not have variables in  table format and pixels do not have distinct semantic meaning as traditional variables. Instead, the learned representations of images can be mapped to traditional variables with semantic meaning. We design this experiment to demonstrate the capability of DIV.VAE with image data.

To simulate this, we replaced the covariates $\mathbf{X}$ with the pixels of two handwritten digits from the MNIST dataset~\cite{lecun2010mnist} as done in the literature~\cite{rissanen2021critical, yuan2022auto}. The dimension of $\mathbf{X}$ is $2*28*28=1568$. The datasets are generated based on the DAG in Fig.~\ref{fig:figure_image} (a), and the specifications are as follows: $U^{'} \sim N(0, 0.05)$, where $N( , )$ denotes the normal distribution. The treatment assignment $W$ is generated from $n$ (where $n$ is the sample size) Bernoulli trials using the assignment probability based on variables $\{\mathbf{Z}, \mathbf{C}\}$ where $\mathbf{Z}$ represents the tens digit of a two-digit handwritten number (i.e., IV), and $\mathbf{C}$ represents the ones digit of a two-digit handwritten number and latent variables $\{U^{'}\}$ as $P(W = 1 \mid \mathbf{Z}, \mathbf{C}, U^{'})= [1+exp\{2- U^{'} - \mathbf{Z}- \mathbf{C}\}]^{-1}$. The outcome is formulated by $Y = 2*\mathbf{C} + 10*W + U^{'} +\epsilon_{y}$. Note that the causal effect of $W$ on $Y$ is fixed at 10.

To demonstrate the representation learning ability of DIV.VAE, we consider an Oracle setting as the baseline. The Oracle set can access the outcome label from image data. $\mathbf{Z}$ and $\mathbf{C}$ are read from a two-digit handwritten number. Note that in this case, bias is not zero because of $\{U^{'}\}$ and $\epsilon_y$. Instead, DIV.VAE learns the representation $\mathbf{Z}$ and $\mathbf{C}$ from $\mathbf{X}$ (i.e., a two-digit handwritten number, by concatenating or combining two images from the MNIST dataset).

\textbf{Results.}  The results are visualised in Fig.~\ref{fig:figure_image} (b) with the mean and STD of 30 runs. The Oracle generates some estimation bias due to inconsistency between the generated datasets and the true causal DAG.  This bias decreases as the sample size increases. DIV.VAE performs worse than Oracle with small sample sizes but improves significantly as the sample size grows. At sample sizes of 10k and 50k, DIV.VAE performs similarly to the Oracle, demonstrating its ability to extract valid IV from high-dimensional image data.

\subsection{Parameters Analysis}
\label{subsec:para}
  With the DIV.VAE algorithm,  two tuning parameters, namely $\alpha_W$ and $\alpha_Y$, are used to balance $\mathcal{L}_{ELBO}$ and the two classifiers. We examine the parameter settings $\alpha_W$ and $\alpha_Y$ across a range of values, specifically $ \{0.01, 0.1, 1, 10, 100, 1,000, 10,000\}$, to analyse the sensitivity of DIV.VAE on synthetic datasets with a sample size of 10k. Note that the package \emph{pyro} requires $\alpha_W$ and $\alpha_Y$ to be the same. These datasets are generated using the same data generation process presented in Section~\ref{subsec:simulationstudy}. The estimation biases of DIV.VAE are reported in Table~\ref{tab:parasets}. From Table~\ref{tab:parasets}, we observe that DIV.VAE achieves the smallest estimation bias when both parameters, $\alpha_W$ and $\alpha_Y$, are set to 100. There is a need to tune parameters  $\alpha_W$ and $\alpha_Y$ in an application.

\begin{table}[t]
	\centering
		\caption{Estimation bias in the different setting and different values of tunning parameters $\alpha_W$ and $\alpha_Y$. }
	\begin{tabular}{ccc}
	\hline
	\multirow{2}{*}{$\{\alpha_W, \alpha_Y\}$} & \multicolumn{2}{c}{Dataset}                     \\ \cline{2-3} 
	& \multicolumn{1}{c|}{Linear}       & Nonlinear    \\ \hline
		0.01  & 80.5 ± 51.17  & 86.65 ± 96.38 \\
		0.1   & 67.25 ± 21.59 & 65.15 ± 40.72 \\
		1     & 52.42 ± 25.97 & 57.2 ± 33.12  \\
		10    & 32.57 ± 22.72 & 25.5 ± 7.85   \\
		100   & 13.88 ± 7.65  & 23.13 ± 9.18  \\
		1,000  & 26.26 ± 13.15 & 25.81 ± 15.06 \\
		10,000 & 35.06 ± 9.99  & 31.82 ± 16.62 \\ \bottomrule
	\end{tabular}
		\label{tab:parasets}
\end{table}

\subsection{Evaluation of  the dimension of latent IV representation}
\label{subsec:dimensionsofIVs}
\begin{figure}[t]
\centering
\includegraphics[scale=0.35]{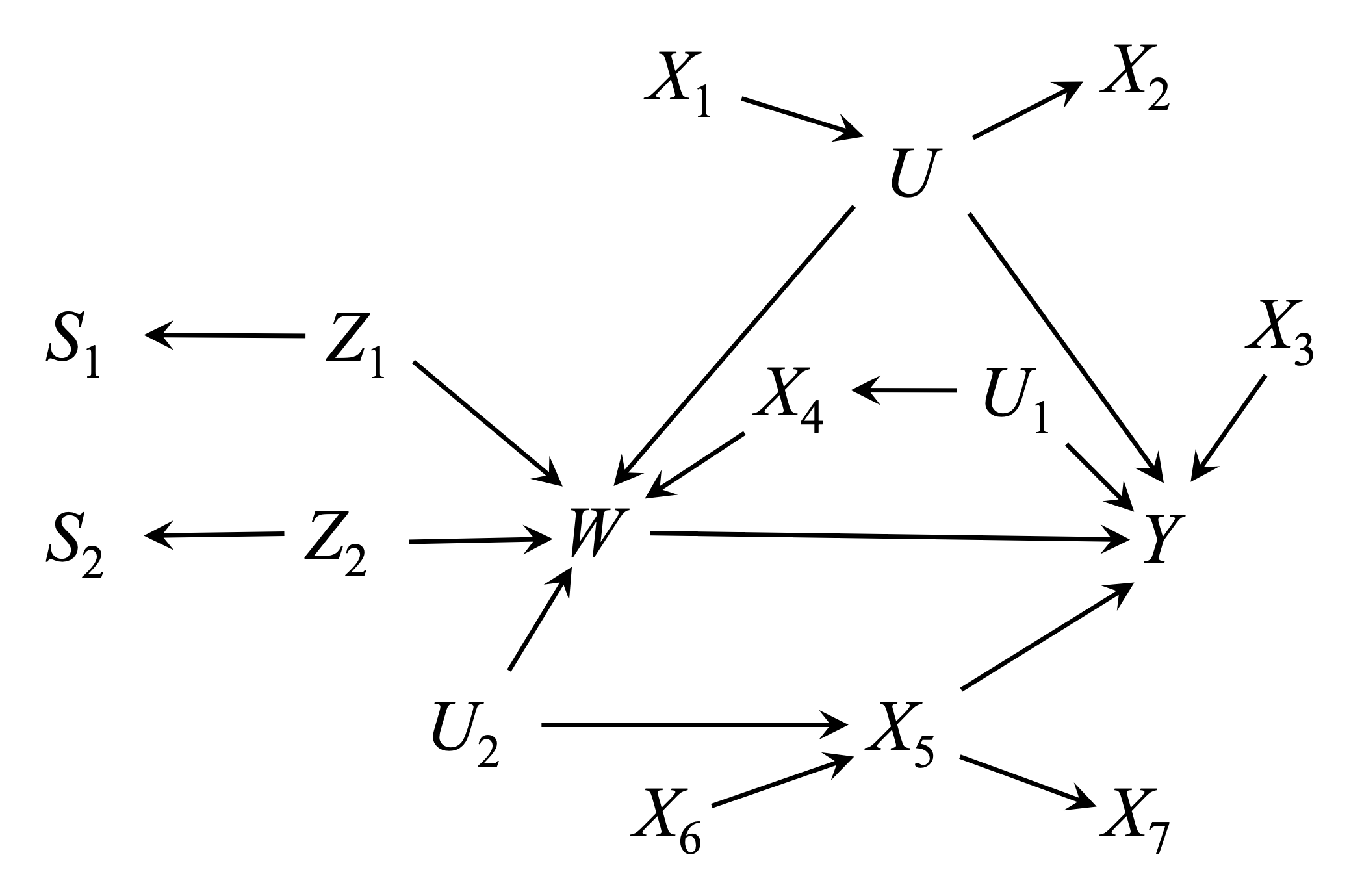}
\caption{The true causal DAG with a latent common cause $U$ between $W$ and $Y$ is used to generate the synthetic datasets.  $\{Z_1, Z_2\}$ and $\{S_1, S_2\}$  are latent IVs and SIVs, respectively.  $\{U_1, U_2\}$ are two latent variables, and other measured variables are pretreatment variables of $(W, Y)$.}
\label{fig:figure_002}
\end{figure}

\begin{figure}[t]
\centering
\includegraphics[scale=0.35]{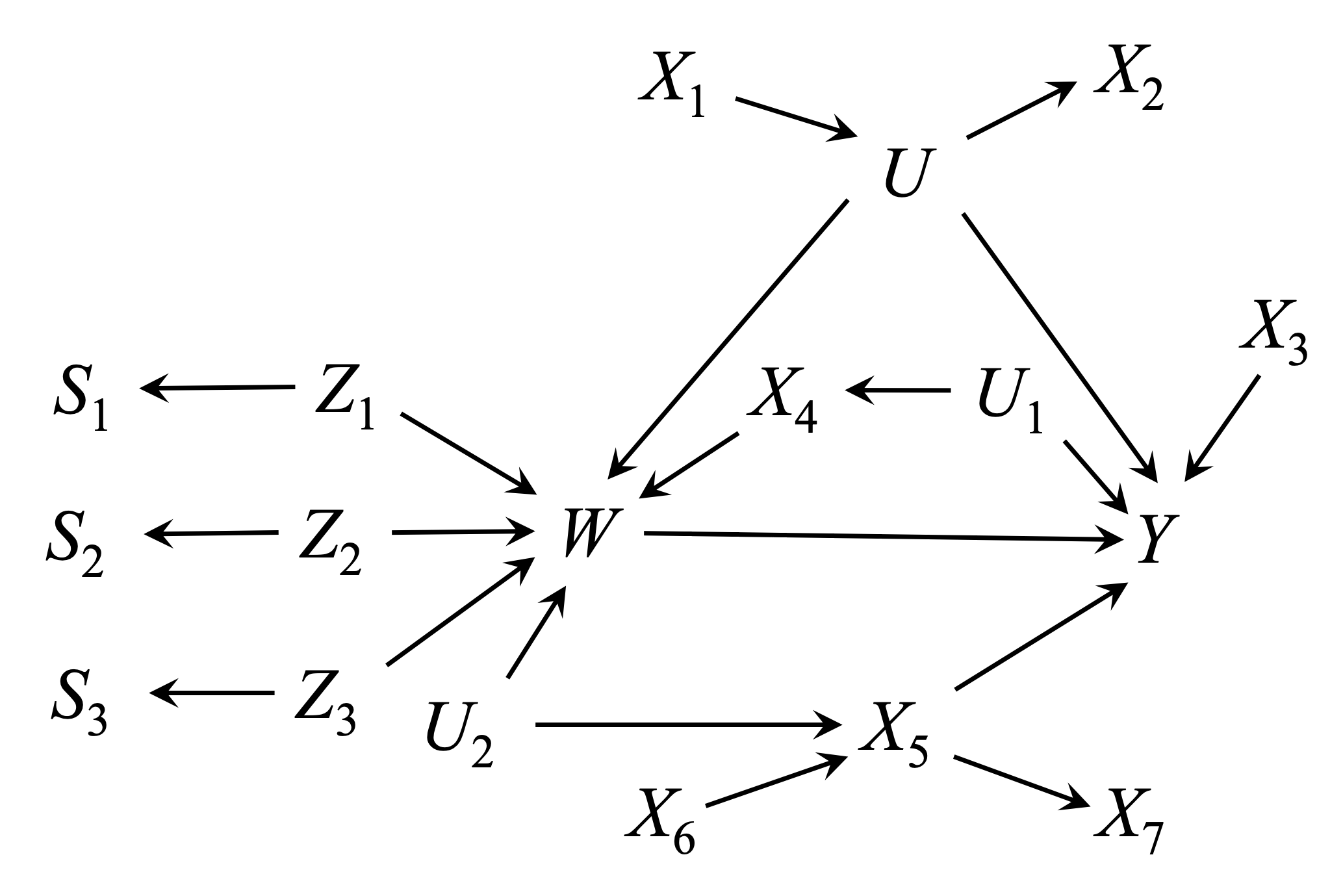}
\caption{The true causal DAG with a latent common cause $U$ between $W$ and $Y$ is used to generate the synthetic datasets. $\{Z_1, Z_2, Z_3\}$ and $\{S_1, S_2, S_3\}$ are latent IVs and  SIVs, respectively.  $\{U_1, U_2\}$ are two latent variables, and other measured variables are pretreatment variables of $(W, Y)$.}
\label{fig:figure_003}
\end{figure}

\begin{figure*}[ht]
\centering
\includegraphics[scale=0.5]{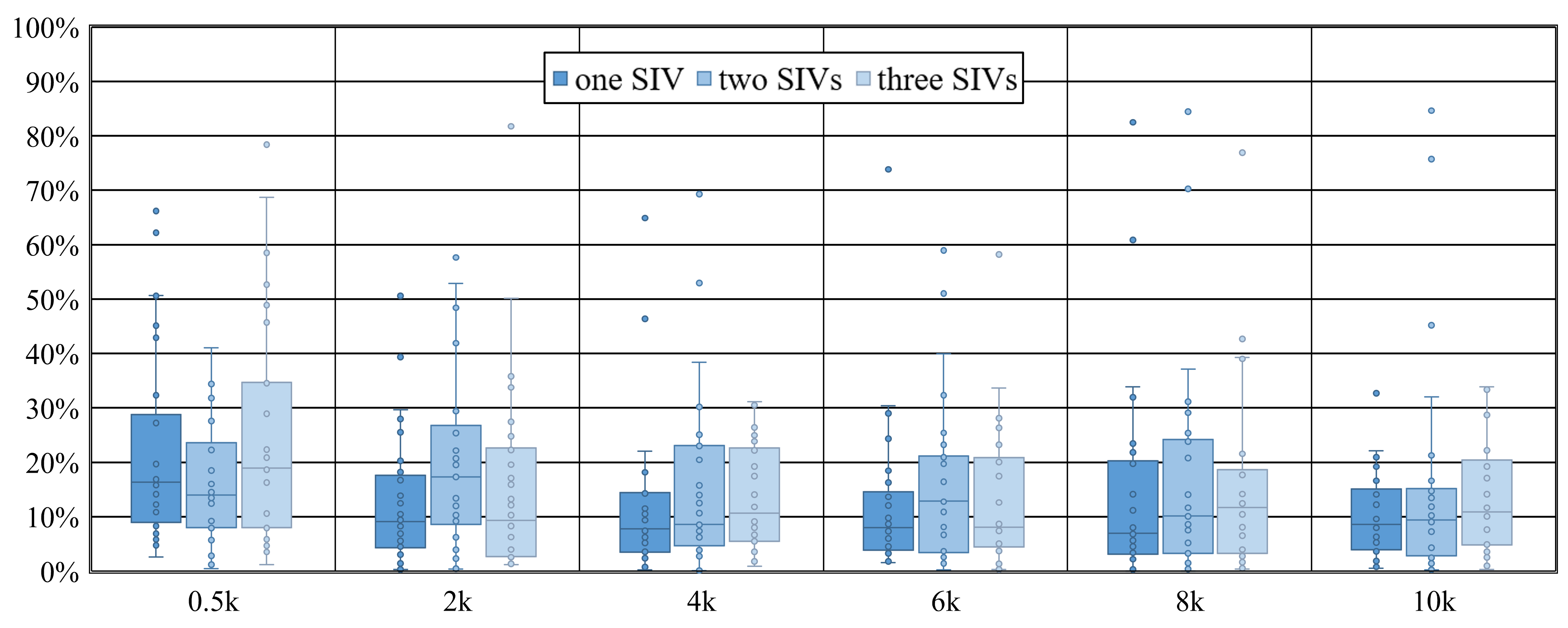}
\caption{The experimental results of DIV.VAE on all three groups of synthetic datasets with $\left|\mathbf{Z}\right|=1$, where the horizontal axis represents the sample size and the vertical axis represents the estimation bias (\%). Mismatching between the number of latent IVs and the number of SIVs does not cause a performance deterioration.} 
\label{fig:thedimensionoflatentIVs}
\end{figure*}

In our implementation, we set $\left|\mathbf{Z}\right|$ to 1 in the disentangling process. We use this experiment to demonstrate the effectiveness of the setting. To do so, we use the same data generation process as used in Section~\ref{subsec:simulationstudy}  with three causal DAGs in Figures ~\ref{fig:figure_001}, ~\ref{fig:figure_002} and~\ref{fig:figure_003} to generate three groups of synthetic datasets such that the three groups of datasets contain one SIV, two SIVs and three SIVs, respectively with other variables and causal relationships remaining unchanged. We repeatedly generate 30 datasets for each group to avoid bias of data generation. The estimation biases of DIV.VAE on the three groups of synthetic data are reported in Fig.~\ref{fig:thedimensionoflatentIVs}.

The second group of datasets are generated from the DAG in Fig.~\ref{fig:figure_002}. The generation processes different from the previous section are discussed in the following: $Z_1, Z_2 \sim N(0, 1);
	\epsilon_{S_1, S_2} \sim N(0, 0.5);
	S_1\sim N(0, 1) + Z_1+\epsilon_{S_1};
	S_2\sim N(0, 1) + Z_2+\epsilon_{S_2}$.

The treatment assignment $W$ is generated from $n$ Bernoulli trials by using the assignment probability $P(W=1\mid U, Z_1, Z_2, X_4, X_5, U_2) = [1+exp\{2-2*U-2*Z_1-2*Z_23*X_4- X_5 -3*U_2\}]^{-1}$. The generation processes of other variables are the same as done in Section~\ref{subsec:simulationstudy}.

The third group of datasets are generated from the DAG in Fig.~\ref{fig:figure_003}. The generation processes different from the  SectionSection~\ref{subsec:simulationstudy} are discussed in the following: 
 $Z_1, Z_2, Z_3  \sim N(0, 1);
\epsilon_{S_1, S_2, S_3} \sim N(0, 0.5);
	S_1\sim N(0, 1) + Z_1+\epsilon_{S_1};
	S_2\sim N(0, 1) + Z_2+\epsilon_{S_2};
S_3\sim N(0, 1) + Z_3+\epsilon_{S_3}$.

The treatment assignment $W$ is generated from $n$ Bernoulli trials by using the assignment probability: $P(W=1\mid U, Z_1, Z_2, Z_3, X_4, X_5, U_2) = [1+exp\{2-2*U-2*Z_1-2*Z_2-2*Z_3+3*X_4- X_5 -3*U_2\}]^{-1}$. The generation processes of other variables are the same as discussed in Section Simulation study.

From Fig.~\ref{fig:thedimensionoflatentIVs}, we see that the estimation biases of DIV.VAE on all three groups of synthetic datasets are consistently small. That means, it is safe to set  $\left|\mathbf{Z}\right|$ to $1$ for DIV.VAE when we have no enough knowledge about the number of SIVs in real-world datasets.

\subsection{Ablation study}
\label{subsec:as}
Here we develop a variant DIV.VAE to explore the component of the Orthogonality Promoting Regularisation (OPR) for the average causal effect estimation from data with latent confounders. The variant of DIV.VAE is the loss function in Eq. (7) in the main text without the OPR term. The variant DIV.VAE is referred to as $\text{DIV.VAE}_\text{w/o.OPR}$.

\begin{figure}[ht]
\centering
\includegraphics[scale=0.49]{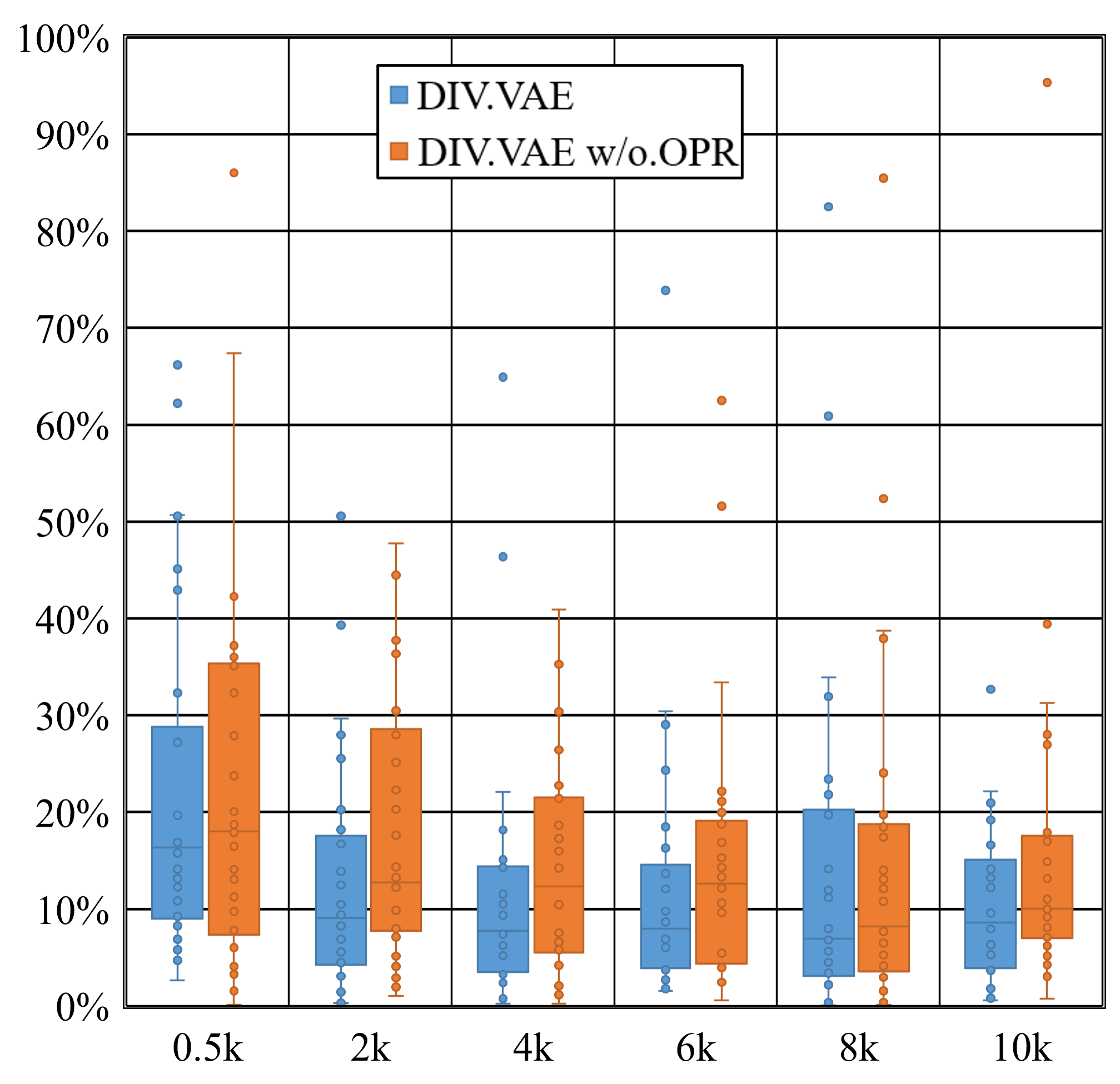}
\caption{Ablation study of DIV.VAE on synthetic datasets, where the horizontal axis represents the sample size and the vertical axis represents the estimation bias (\%).}
\label{fig:ablation_study}
\end{figure}

We conduct the ablation study experiment on the synthetic datasets that are used in the Section Simulation study of the main text to show the performance of the proposed DIV.VAE  with  $\text{DIV.VAE}_\text{w/o.OPR}$. The experimental results are visualised in Fig.~\ref{fig:ablation_study}.

\textbf{Results}. From the results in Fig.~\ref{fig:ablation_study},  $\text{DIV.VAE}_\text{w/o.OPR}$ has a larger estimation bias than DIV.VAE on all synthetic datasets. Moreover, the variance of $\text{DIV.VAE}_\text{w/o.OPR}$ is also larger than DIV.VAE. The observations show the importance of the OPR in learning and disentangling the latent IV representation $\mathbf{Z}$ from the latent representation $\mathbf{\Phi}=(\mathbf{Z, C})$ for causal effect estimation from data with latent confounders. 

\subsection{The empirical evaluation on the independence relation of $\mathbf{Z}$ and $\mathbf{C}$}
\label{subsec:zc}
In this section, we conduct an empirical evaluation of the independence relation of $\mathbf{Z}$ and $\mathbf{C}$ by using the synthetic datasets generated in  Section `Simulation study'. In the empirical evaluation, we use the Pearson product-moment correlation coefficients (PCCs) as the evaluation index. In general, two variables are unrelated when PCC is less than 0.3. We report the mean with the standard deviation (std) over 30 replications. In all experiments, $\left|\mathbf{Z}\right| =1$ and $\left|\mathbf{C}\right| =10$. Hence, we calculate the PCC of each pair $(Z, C_i)$ where $C_i\in \mathbf{C}$. The PCCs of each pair $(Z, C_i)$ in $(\mathbf{Z}, \mathbf{C})$ by using $\text{DIV.VAE}_\text{w/o.OPR}$ and DIV.VAE are reported in Table~\ref{tab:divvaewoOPR} and Table~\ref{tab:divvae}, respectively.

\begin{table*}[t]
\centering
\caption{The PCCs of each pair of $(Z, C_i)$ for $\text{DIV.VAE}_\text{w/o.OPR}$ (mean ± std).}
\begin{tabular}{ccccccc}
	\toprule
	N       & 0.5k          & 2k        & 4k        & 6k        & 8k         & 10k        \\ \midrule
	PCC($Z$,$C_1$)   & 0.246 ± 0.15 & 0.232 ± 0.15 & 0.141 ± 0.09 & 0.174 ± 0.11 & 0.107 ± 0.09 & 0.107 ± 0.07 \\
	PCC($Z$,$C_2$)   & 0.263 ± 0.15 & 0.209 ± 0.14 & 0.189 ± 0.12 & 0.150 ± 0.10 & 0.159 ± 0.09 & 0.092 ± 0.08 \\
	PCC($Z$,$C_3$)   & 0.256 ± 0.16 & 0.201 ± 0.13 & 0.175 ± 0.10 & 0.143 ± 0.12 & 0.133 ± 0.10 & 0.101 ± 0.07 \\
	PCC($Z$,$C_4$)   & 0.307 ± 0.15 & 0.208 ± 0.14 & 0.193 ± 0.14 & 0.185 ± 0.10 & 0.143 ± 0.10 & 0.123 ± 0.08 \\
	PCC($Z$,$C_5$)   & 0.277 ± 0.17 & 0.196 ± 0.12 & 0.166 ± 0.12 & 0.166 ± 0.12 & 0.141 ± 0.11 & 0.091 ± 0.07 \\
	PCC($Z$,$C_6$)   & 0.237 ± 0.16 & 0.218 ± 0.14 & 0.189 ± 0.13 & 0.187 ± 0.10 & 0.144 ± 0.12 & 0.097 ± 0.08 \\
	PCC($Z$,$C_7$)   & 0.243 ± 0.15 & 0.181 ± 0.14 & 0.149 ± 0.11 & 0.147 ± 0.10 & 0.110 ± 0.09 & 0.086 ± 0.06 \\
	PCC($Z$,$C_8$)   & 0.254 ± 0.17 & 0.185 ± 0.12 & 0.173 ± 0.11 & 0.136 ± 0.09 & 0.125 ± 0.08 & 0.093 ± 0.07 \\
	PCC($Z$,$C_9$)   & 0.267 ± 0.17 & 0.230 ± 0.14 & 0.160 ± 0.10 & 0.147 ± 0.11 & 0.139 ± 0.10 & 0.103 ± 0.10 \\
	PCC($Z$,$C_{10}$)   & 0.293 ± 0.12 & 0.207 ± 0.15 & 0.174 ± 0.13 & 0.164 ± 0.13 & 0.142 ± 0.11 & 0.101 ± 0.09 \\ \midrule
	Average & 0.264 ± 0.16 & 0.207 ± 0.14 & 0.171 ± 0.11 & 0.160 ± 0.11 & 0.134 ± 0.10 & 0.101 ± 0.08 \\ \bottomrule
\end{tabular}
\label{tab:divvaewoOPR}
\end{table*}

\begin{table*}[t]
\centering
\caption{The PCCs of each pair of $(Z, C_i)$  for DIV.VAE (mean ± std).}
\begin{tabular}{ccccccc}
	\toprule
	N       & 0.5k          & 2k        & 4k        & 6k        & 8k         & 10k        \\ \midrule
	PCC($Z$,$C_1$)   & 0.206 ± 0.13 & 0.177 ± 0.11 & 0.142 ± 0.11 & 0.092 ± 0.08 & 0.082 ± 0.08 & 0.055 ± 0.04 \\
	PCC($Z$,$C_2$)   & 0.203 ± 0.14 & 0.172 ± 0.11 & 0.100 ± 0.08 & 0.094 ± 0.06 & 0.080 ± 0.06 & 0.062 ± 0.04 \\
	PCC($Z$,$C_3$)   & 0.180 ± 0.14 & 0.205 ± 0.10 & 0.152 ± 0.09 & 0.104 ± 0.10 & 0.080 ± 0.05 & 0.052 ± 0.04 \\
	PCC($Z$,$C_4$)   & 0.225 ± 0.14 & 0.177 ± 0.09 & 0.083 ± 0.09 & 0.081 ± 0.06 & 0.064 ± 0.06 & 0.089 ± 0.06 \\
	PCC($Z$,$C_5$)   & 0.194 ± 0.13 & 0.152 ± 0.11 & 0.113 ± 0.08 & 0.088 ± 0.06 & 0.080 ± 0.07 & 0.070 ± 0.07 \\
	PCC($Z$,$C_6$)   & 0.199 ± 0.11 & 0.164 ± 0.11 & 0.141 ± 0.11 & 0.101 ± 0.08 & 0.071 ± 0.05 & 0.089 ± 0.07 \\
	PCC($Z$,$C_7$)   & 0.165 ± 0.11 & 0.189 ± 0.11 & 0.168 ± 0.10 & 0.090 ± 0.08 & 0.097 ± 0.07 & 0.071 ± 0.06 \\
	PCC($Z$,$C_8$)   & 0.207 ± 0.11 & 0.153 ± 0.13 & 0.167 ± 0.11 & 0.071 ± 0.06 & 0.097 ± 0.08 & 0.081 ± 0.07 \\
	PCC($Z$,$C_9$)   & 0.210 ± 0.13 & 0.199 ± 0.13 & 0.118 ± 0.09 & 0.069 ± 0.05 & 0.071 ± 0.05 & 0.075 ± 0.06 \\
	PCC($Z$,$C_{10}$)   & 0.192 ± 0.12 & 0.205 ± 0.14 & 0.137 ± 0.11 & 0.106 ± 0.08 & 0.079 ± 0.08 & 0.070 ± 0.06 \\ \midrule
	Average & 0.198 ± 0.13 & 0.179 ± 0.11 & 0.132 ± 0.10 & 0.089 ± 0.07 & 0.080 ± 0.06 & 0.071 ± 0.06 \\ \bottomrule
\end{tabular}
\label{tab:divvae}
\end{table*}

\textbf{Results}. From both Tables, we have three observations: (1) The PCC between $Z$ and each $C_i$ in $\mathbf{C}$ is less than 0.3, \ie $Z$ and each $C_i$ in $\mathbf{C}$ are uncorrelated. (2) As the sample increases, the mean of the PCCs between $Z$ and each $C_i$ in $\mathbf{C}$ dropped significantly. (3)  As the sample increases, the std of the PCCs between $Z$ and each $C_i$ in $\mathbf{C}$ are also decreased. 

Therefore, these results show that $Z$ and each $C_i$ in $\mathbf{C}$ are well disentangled.  

By comparing Table~\ref{tab:divvaewoOPR} and Table~\ref{tab:divvae}, we know that the PCCs of DIV.VAE is significantly smaller than the PCCs of $\text{DIV.VAE}_\text{w/o.OPR}$'s on all synthetic datasets. It further confirms that the OPR term plays an important role in encouraging $\mathbf{Z}\ci\mathbf{C}$ in learning and disentangling the latent IV representation $\mathbf{Z}$ from the latent representation $\mathbf{\Phi}=(\mathbf{Z, C})$ of $\mathbf{X}$.

\section{Related work}
\label{sec:relwork}
In this section, we review the research closely related to our work, including IV based methods with a given IV, data-driven IV based methods without a given IV and deep learning (including VAE) based causal effect estimation.	

\textbf{IV based methods with a given IV}. In practice, before we use an IV method, one needs to nominate a valid IV based on  domain knowledge. Under the assumption that a valid IV has been given, several IV based counterfactual prediction methods have been developed for heterogeneous causal effect estimation, such as instrumental random forest regression~\cite{athey2019generalized}, generalised method of moments (GMM)~\cite{bennett2019deep}, deep ensemble method for the instrumental variable (DeepIV)~\cite{hartford2017deep} and kernel instrumental variable regression (KIV)~\cite{singh2019kernel}. Yuan \etal \cite{yuan2022auto} proposed a novel AutoIV algorithm to automatically generate IV representation for the downstream IV based counterfactual prediction under the assumption that the latent confounder between $W$ and $Y$ is independent of the set of measured covariates, but this assumption may be violated in many real applications.  Furthermore, AutoIV requires $S\ci Y\mid W$, which is more strict than DIV.VAE. Hence, AutoIV does not solve the same problem as DIV.VAE does and is not compared in the experiments.  Different from the above mentioned IV methods, we focus on learning a valid IV representation from data without nominating a valid IV from domain knowledge. 

\textbf{Data-driven IV based method without a given IV}. In the absence of a given IV, a few data-driven methods have been proposed for finding valid IVs~\cite{silva2017learning} or synthesising IVs~\cite{burgess2013use,kuang2020ivy}  or eliminating the effect of invalid IVs by using statistical analysis~\cite{kang2016instrumental,guo2018confidence,hartford2021valid}. For example, IV.Tetrad~\cite{silva2017learning} uses the tetrad constraint to perform statistical tests for discovering pairs of valid IVs. Kuang \etal \cite{kuang2020ivy} proposed the Ivy method to combine IV candidates as a summary IV for identifying all invalid IVs or dependencies. Kang \etal \cite{kang2016instrumental} proposed the sisVIVE method to estimate causal effects when at least  half of the covariates are valid IVs (\ie majority assumption). Hartford \etal \cite{hartford2021valid} developed a ModeIV algorithm by employing a deep learning based IV estimator~\cite{hartford2017deep} under the majority assumption. Both sisVIVE and ModeIV rely on the majority assumption, but it is difficult to verify the majority assumption and this limits their applications. Unlike this type of data-driven method, DIV.VAE only needs an SIV and makes use of VAE to recover IV information. 

\textbf{Deep learning (including VAE) based causal effect estimation}. 
The existing VAE-based causal effect estimators~\cite{louizos2017causal,hassanpour2019learning,zhang2021treatment} simply assume no latent confounders, and thus they are not for dealing with latent confounders. Moreover, this type of VAE-based estimators relies on an impractical assumption~\cite{rissanen2021critical}, \ie they require all covariates to be measured as the proxy variables of the latent confounders or latent representations~\cite{louizos2017causal}. Under the unconfoundedness assumption, researchers focus on designing deep learning models for estimating causal effects from observational data, \eg Balance Learning Representation neutral network (BLR)~\cite{johansson2016learning}, Counterfactual Regression (CFR)~\cite{shalit2017estimating}, and Generative Adversarial Nets for estimating Individualised Treatment Effects (GANITE)~\cite{yoon2018ganite}. However, none of these deep learning based estimators is able to obtain an unbiased estimation of the causal effect of $W$ on $Y$ in the presence of a latent confounder between $(W, Y)$. So our work is the first one to use the VAE model in learning and disentangling the latent IV representation from the latent representation of the measured covariates for causal effect estimation from data without the unconfoundedness assumption. 

When there is a latent confounder between $(W, Y)$, the causal effect of $W$ on $Y$ is non-identifiable with covariate adjustment~\cite{pearl2009causality,cheng2022data} and these deep learning methods do not work as they are based on covariate adjustment. Our method takes the IV approach, a practical way to address this challenging problem very well.

\section{Conclusion}
\label{sec:con}
Causal effect estimation from data with latent variables is crucial for many real-world applications, but there is a lack of effective data-driven methods for dealing with latent confounders. In this work, we make a connection between surrogate instrumental variables studied in the causal inference and statistics communities and the VAE model widely used in the machine learning community for latent representation learning. This connection has provided the theoretical guarantees for us to develop the DIV.VAE method to learn the latent IV representation through VAE-based disentangled representation learning. This in turn enables us to leverage the IV approach to obtain unbiased causal effect estimation from data in the presence of latent confounders. To the best of our knowledge, this is the first work to establish a link between generative modelling and the IV approach. Extensive experiments on synthetic and real-world datasets demonstrate that DIV.VAE is very effective in estimating the average causal effect from data with latent variables. We believe that the findings presented in this work have the potential to significantly improve the real-world applications of the IV approach for inferring unbiased causal effects from data with latent variables.

In our future work, we plan to explore connections between balanced representation learning~\cite{shalit2017estimating,imbens2015causal}, proximal causal learning~\cite{miao2018identifying,tchetgen2020introduction} and latent IV representation learning using the disentanglement technique presented in this work. Furthermore, we will extend our DIV.VAE to applications in recommendation systems~\cite{wang2020causal}, natural language processing~\cite{feder2022causal}, and other areas~\cite{scholkopf2021toward}.

\section*{Acknowledgments}
We thank the action editors and the reviewers for their invaluable comments and suggestions. We wish to acknowledge the support from the Australian Research Council (under grant DP230101122).

\bibliographystyle{IEEEtran}
\bibliography{DIVVAE.bib}

\end{document}